%% file: ManiF-KDD26.tex
\documentclass[sigconf,nonacm]{acmart}

\usepackage{amsmath,amsfonts}

\usepackage{amsthm}

\newtheorem{proposition}{\textbf{Proposition}}

\usepackage{float}

\usepackage{color}
\usepackage{tikz}
\usetikzlibrary{trees}
\usepackage{amsthm}
\usepackage{makecell}
\usepackage{tikz}
\usetikzlibrary{trees}
\usepackage{tabularx,colortbl}
\usepackage{threeparttable}
\usepackage{booktabs}
\usepackage{multirow}
\usepackage{subfig}
\usepackage{graphicx}
\usepackage{url}
\usepackage{enumitem}
\usepackage{tcolorbox}
\usepackage{xcolor}
\usepackage[ruled, vlined, linesnumbered]{algorithm2e}
\usepackage{hyperref}
\usepackage[noabbrev]{cleveref}
\crefname{equation}{Eq.}{Eqs.}
\usepackage{soul}
\usepackage{adjustbox}

\definecolor{DeepPink}{HTML}{FF1493}
\definecolor{Orchid}{HTML}{DA70D6}
\definecolor{Magenta}{HTML}{FF00FF}
\definecolor{Fuchsia}{HTML}{FF00FF}
\definecolor{LavenderPink}{HTML}{FFB6C1}
\definecolor{verylightgray}{rgb}{0.9, 0.9, 0.9}
\definecolor{lightred}{rgb}{1,0.8,0.8}

\AtBeginDocument{%
  }

\setcopyright{acmlicensed}
\copyrightyear{2026}
\acmYear{2026}
\acmDOI{XXXXXXX.XXXXXXX}
\acmConference[Conference acronym 'XX]{Make sure to enter the correct
  conference title from your rights confirmation email}{June 03--05,
  2018}{Woodstock, NY}
\acmISBN{978-1-4503-XXXX-X/2018/06}

\begin{document}

\title{Approximate Machine Unlearning through Manifold Representation Forgetting Guided by Self Mode Connectivity}
\subtitle{ \it \textbf{ To appear at KDD 2026. Author version} }

\author{Weiqi Wang}
\email{weiqi.wang@xjtu.edu.cn}
\orcid{0000-0002-7905-3126}
\affiliation{%
	\institution{Xi'an Jiaotong University}
	\city{Xi'an}
	\state{Shaanxi}
	\country{China}}

\author{Zhiyi Tian}
\email{zhiyi.tian@ieee.org}
\orcid{0000-0001-8905-0941}
\affiliation{%
	\institution{Southeast University}
	\city{Nanjing}
	\state{Jiangsu}
	\country{China}
}

\author{Chenhan Zhang}
\email{chzhang@ieee.org}
\orcid{0000-0001-8905-0941}
\affiliation{%
	\institution{University of Technology Sydney}
	\city{Sydney}
	\state{NSW}
	\country{Australia}
}

\author{Luoyu Chen}
\email{luoyu.chen@uts.edu.au}
\orcid{0000-0001-8905-0941}
\affiliation{%
	\institution{University of Technology Sydney}
	\city{Sydney}
	\state{NSW}
	\country{Australia}
}

\author{Shui Yu}
\email{shui.yu@uts.edu.au}
\orcid{0000-0003-4485-6743}
\affiliation{%
	\institution{University of Technology Sydney}
	\city{Sydney}
	\state{NSW}
	\country{Australia}
}

\renewcommand{\shortauthors}{WeiqiWang, Zhiyi Tian, Chenhan Zhang, Luoyu Chen, and Shui Yu}

\input{Contents/0_abstract}



\maketitle

 \input{Contents/1_intro}

\input{Contents/2_related_work}

\input{Contents/3_problem_definition}

\input{Contents/4_approach}

\input{Contents/5_experiments}

\input{Contents/6_summary}



\bibliographystyle{ACM-Reference-Format}
\bibliography{mu_mcr}

\newpage
\appendix

\input{Contents/Appendix/appendix1}

\input{Contents/Appendix/appendix2}
\end{document}

%% file: Contents/0_abstract.tex
\begin{abstract}

Machine unlearning is a fundamental mechanism that enforces the right to be forgotten. Existing unlearning studies that rely on label manipulation or task-gradient reversal often deliver limited unlearning effectiveness. Moreover, they can undermine the original learning objective and typically do not guarantee equivalence to standard unlearning by retraining. 

In this paper, we propose \textbf{ManiF-SMC} (\textbf{Mani}fold \textbf{F}orgetting with \textbf{S}elf \textbf{M}ode \textbf{C}onnectivity), motivated by the observation that a model retrained on the remaining data tends to classify erased samples by their semantic similarity to the retained data. We begin with systematically recasting the approximate unlearning as pushing each erased sample away from its original learned manifold representation centroid toward its nearest semantic neighbors in the retained data. This reformulation aligns unlearning with retraining behavior and operates purely in representation space, reducing reliance on labels and task-specific gradients. To tackle the manifold representation-based unlearning problem, ManiF-SMC encapsulates the unlearning and representation preservation goals in a margin-based triplet loss. Because finding a suitable margin for unlearning is challenging, we propose a self-mode-connectivity module that rapidly reconstructs the local manifold to guide the adaptive margins generation for each unlearning case. Extensive experiments on four representative datasets show that ManiF-SMC achieves unlearning effectiveness comparable to state-of-the-art approximate methods while operating solely within the model's representation space. 


\end{abstract}

%% file: Contents/1_intro.tex
\section{Introduction} \label{intro}

Heightened awareness around data privacy has ushered in rigorous regulatory efforts, exemplified by the legislated laws such as the General Data Protection Regulation (GDPR)~\citep{mantelero2013eu}. These legal frameworks guarantee individuals the right to be forgotten, sparking a nascent area that focuses on removing the influence of specified samples from trained machine learning (ML) models, i.e., machine unlearning~\citep{bourtoule2021machine,warnecke2024machine}. Although retraining from scratch offers the most faithful approach to unlearning, its computational overhead is often prohibitive, spurring the development of more practical approximate unlearning methods~\citep{nguyen2020variational,guo2019certified}.

\begin{figure}[t]
	\centering
	\hspace{-2mm}
	\subfloat{		\label{fig:garandomforgetting10pcifar10400}
		\includegraphics[scale=0.242]{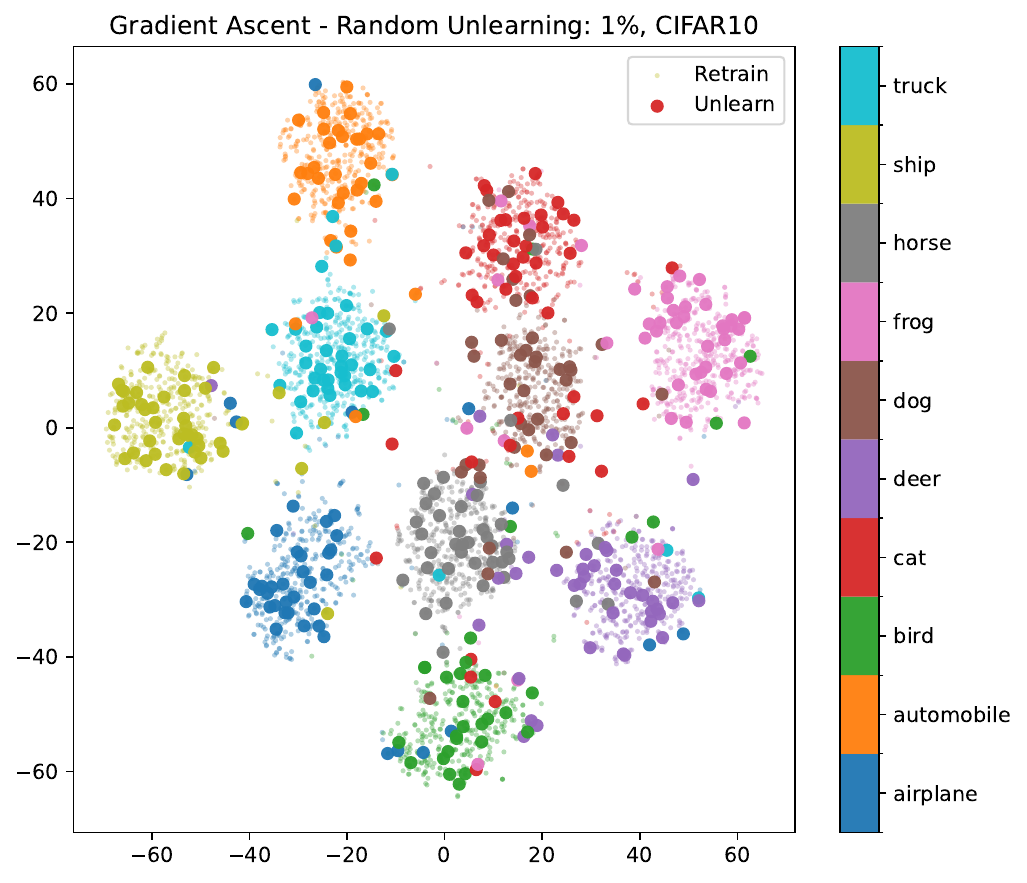}
	}			
	\hspace{-4mm}
	\subfloat{  	\label{fig:retrainrandomforgetting10pcifar10400}
		\includegraphics[scale=0.242]{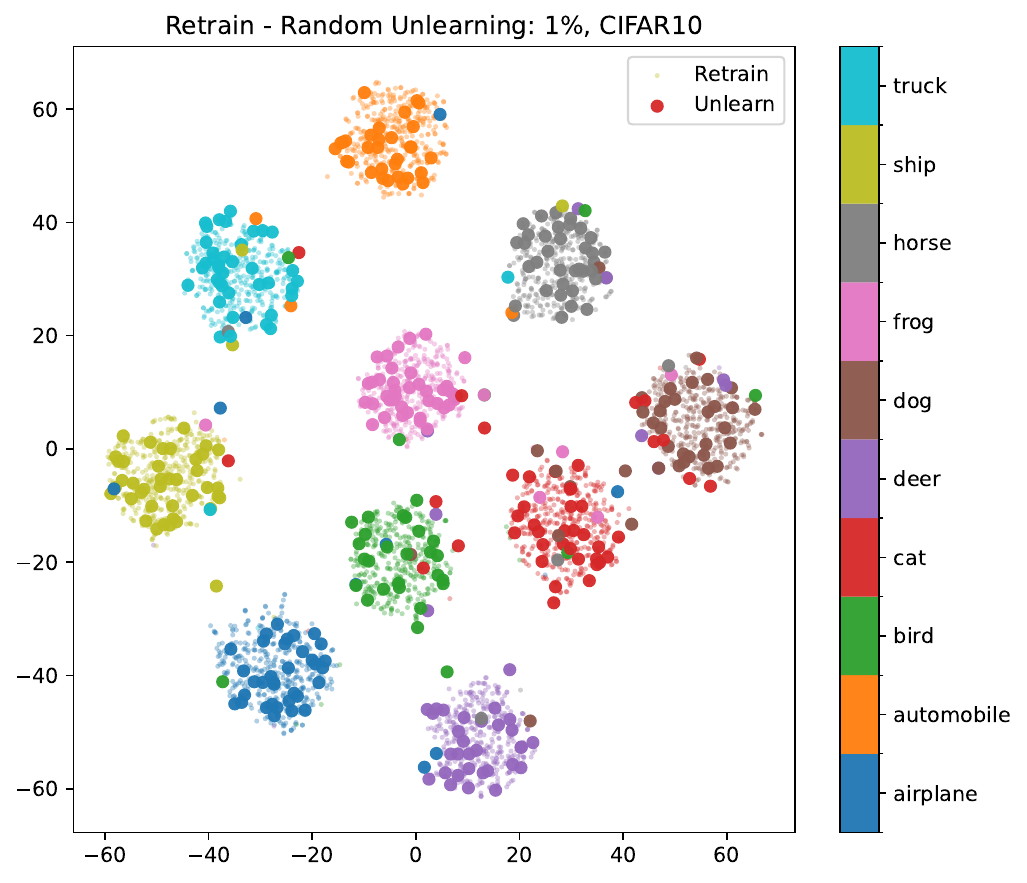}
	}	
	\caption{Representation space of the Gradient Ascent unlearned model (\textit{left}) and Retriened model (\textit{right}) on CIFAR10 after unlearning 1\% (500) randomly selected samples. Small points denote retained samples from different clusters. Larger points denote forgotten samples, assigned to the retained cluster with the highest semantic similarity. 
		\vspace{-4mm}
	} 
	\label{representation_retrain_observation} 
\end{figure}
 
Most existing approximate unlearning methods are considered from a data-centric perspective, requiring the manipulation of labels or the reversal of task-relevant gradients, improving efficiency but limiting unlearning effectiveness. In particular, we face two main challenges. Firstly, unlearning based on label manipulation and gradient reversal conflicts with the original learning objective and typically does not guarantee equivalence to gold-standard unlearning by retraining from scratch \citep{thudi2022necessity, ebrahimpourboroojeny2025not}. Secondly, most approximate unlearning methods construct the unlearning loss heavily relying on the label access to manipulate the label or reverse the task-specific gradients \citep{nguyen2020variational,neel2021descent}. However, in some complex scenarios, such as semantic communication \citep{huang2022toward,tian2024rose} and federated learning \citep{li2020federated}, the unlearning server may only access the model or learned representations. Based on these two challenges, we ask the following question.

\noindent
\textbf{Research Question.} \ul{How can we achieve effective approximate unlearning in representation space, reducing reliance on labels, while remaining consistent with retraining-from-scratch unlearning?}


\noindent
\textbf{Inspiration.}
To develop an effective approximate unlearning method, we first examine how a retraining-from-scratch model represents the erased samples at test time. 
\Cref{representation_retrain_observation} compares the representation space of Gradient Ascent unlearning \citep{graves2021amnesiac,thudi2022unrolling} and retraining from scratch when forgetting 1\% randomly selected CIFAR10 samples. We observe that, under retraining from scratch, erased samples tend to be mapped close to the retained samples that are most semantically similar to them. This motivates a representation-space reformulation of approximate unlearning: we encourage erased samples to move toward the region of their most similar retained samples, so that the resulting model behavior better matches retraining-from-scratch unlearning.

Moreover, we noticed that learned representations have been extensively studied \citep{yerxa2023learning,wang2020understanding}. In self-supervised learning, promoting compactness and uniformity is known to improve representation quality and downstream performance \citep{tian2021understanding}. Building on this line, the maximum manifold capacity representations (MMCR) framework \citep{yerxa2023learning} shows that regulating representation geometry increases class separability and supports high-quality recognition. These results suggest that controlling representation structure can steer model behavior without relying on label manipulation or task-specific gradients. Therefore, we investigate approximate unlearning from a representation-space perspective and use MMCR as a principled regularization tool.

\begin{figure}[t]
	\centering
	\includegraphics[width=0.97\linewidth]{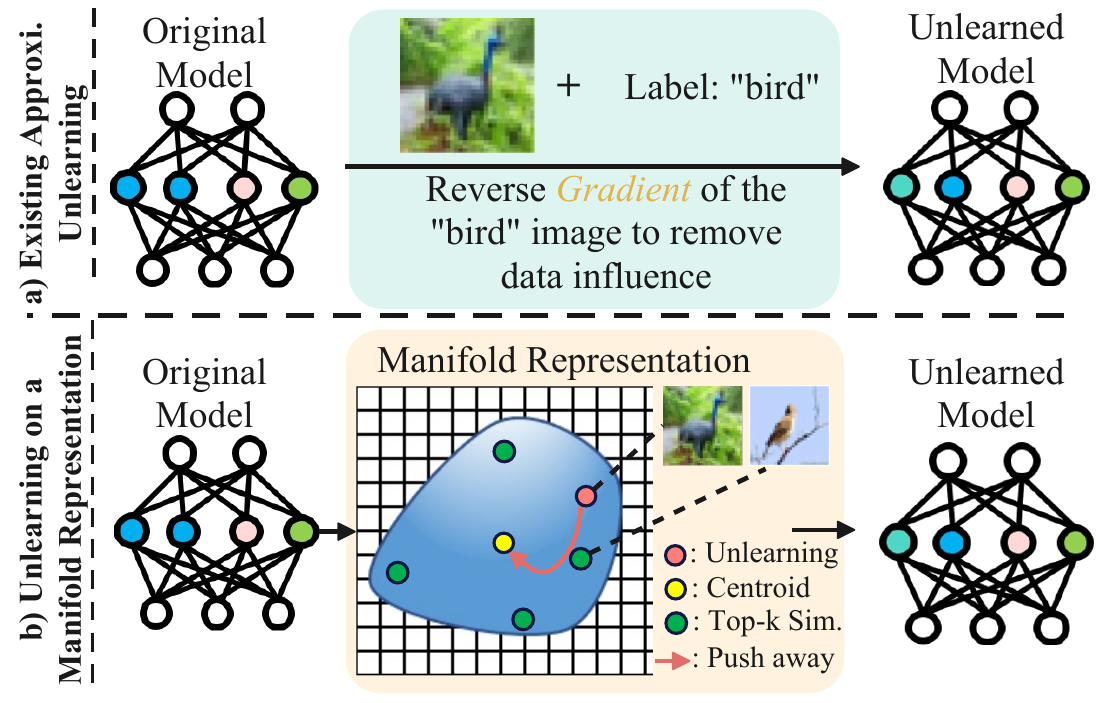} 
	\caption{Comparison with existing approximate unlearning studies. (a) Existing approximate unlearning studies utilize the data and label information, unlearning by reversing the corresponding gradients. (b) Manifold representation-based unlearning tries to push away the unlearning sample from its original representation towards the centroid of the most semantic similar samples in the retained set.
	}
	\label{fig_figure1}
\end{figure}

\noindent
\textbf{Our Work.}
In this paper, we begin by systematically revisiting and reformulating the approximate unlearning problem through the lens of manifold representations and the observation of retraining results in \Cref{representation_retrain_observation}. Specifically, we interpret approximate unlearning as \emph{moving an erased sample’s representation away from its original location and toward the centroid of its most semantically similar retained samples, from a manifold representation perspective}. 
This reformulation aligns approximate unlearning with retraining behavior and operates purely in representation space, decoupling reliance on labels and task-specific gradients. 
We compare existing approximate unlearning formulations with ours in \Cref{fig_figure1}.

To solve this representation-based unlearning problem, we propose \textbf{Mani}fold \textbf{F}orgetting with \textbf{S}elf \textbf{M}ode \textbf{C}onnectivity (\textbf{ManiF-SMC}). 
ManiF-SMC uses a triplet contrastive unlearning loss: it pushes the erased sample away from its original learned representation while pulling it toward the centroid of similar retained representations, with the separation controlled by a margin. However, both the target centroid (of similar retained samples) and an appropriate margin are hard to determine a priori. We introduce a self mode connectivity module that quickly reconstructs the local manifold structure of these similar retained samples, and uses it to estimate the centroid and adapt the margin during unlearning optimization.

We have two key findings through extensive experiments across model architectures and datasets. First, ManiF-SMC cuts the reliance on task labels by operating purely in representation space, while achieving strong unlearning effectiveness compared with state-of-the-art approximate unlearning baselines. This label-agnostic unlearning formulation also enables deployment in broader settings such as semantic communication systems \citep{huang2022toward,tian2024rose}, where the encoder may serve multiple downstream tasks and labels may be unavailable at the unlearning stage. Second, we observed that adopting the MMCRs for model training can improve class separability and boost unlearning effectiveness for both existing approximate and exact unlearning methods.

Our contributions are summarized as follows.
\begin{itemize}[itemsep=0pt, parsep=0pt, leftmargin=*]
	\item We revisit approximate unlearning from a representation perspective and reformulate it as moving an erased sample’s representation away from its original position and toward the centroid of its most semantically similar retained samples.
	\item We propose \textbf{Mani}fold \textbf{F}orgetting with \textbf{S}elf \textbf{M}ode \textbf{C}onnectivity (\textbf{ManiF-SMC}), a label-agnostic unlearning approach that operates purely in representation space and adaptively estimates the centroid and margin for optimization.
	\item Extensive experiments across datasets and architectures show that ManiF-SMC achieves strong approximate unlearning effectiveness without accessing task labels. We further demonstrate that MMCR-based representation regularization improves class separability and consistently boosts the unlearning performance of both approximate and retraining-based methods. Our code is available at \url{https://anonymous.4open.science/r/ManiF-C120}.
\end{itemize}




%% file: Contents/2_related_work.tex
\section{Related Work} \label{related_w}


Machine unlearning techniques are motivated by the growing privacy concerns of individuals and the corresponding privacy regulations~\cite{cao2015towards,bourtoule2021machine}. The most legitimate approach is retraining from scratch~\cite{cao2015towards,thudi2022necessity}. However, this method is often impractical due to the significant computational and storage costs involved, especially for complex deep-learning tasks. Consequently, numerous studies have sought to develop efficient unlearning solutions~\cite{yanarcane2022unlearning,warnecke2024machine}.

\noindent
\textbf{Exact Unlearning.} 
The exact unlearning methods aim to reduce the computational cost of retraining a new model by redesigning the learning algorithms and storing the intermediate parameters during the learning process \cite{cao2015towards,bourtoule2021machine,yanarcane2022unlearning,wu2020deltagrad}. One popular exact unlearning is introduced in  \cite{bourtoule2021machine}, where the ML server divided the full data into shards and trained sub-models separately in each shard. When unlearning, the server simply needs to remove the erasing data from the corresponding shard and retrain the sub-model of this shard~\cite{bourtoule2021machine}. Exact unlearning has the advantage of completely removing the influence of the unlearned data on the model. However, these methods sacrifice the huge storage space and are inefficient when removal requests are frequent. 

\noindent
\textbf{Approximate Unlearning.}
The approximate unlearning methods aim to implement unlearning using only the original model and the samples to be erased, approximating a model as if it was retrained on the remaining dataset~\cite{nguyen2020variational,shen2024labelagnostic,warnecke2024machine}. One representative approximate unlearning solution is based on the Hessian matrix and Newton updates \cite{warnecke2024machine,sekhari2021remember}, which ensures that a model from which data is erased cannot be distinguished from a model that never observed the data. Another representative solutions are based on the Bayesian inference \cite{nguyen2020variational,fu2022knowledge,nguyen2022markov}, which unlearn an approximate posterior only based on the erased samples. 

Moreover, a line of approximate unlearning methods performs post-hoc model editing by manipulating learned representations or decision boundaries. \citet{fan2024salun} shifts attention from updating the full network to modifying a targeted subset of weights, improving both effectiveness and efficiency. \citet{wang2023machine} proposes representation forgetting by operating on a learned bottleneck representation, while boundary unlearning \citep{chen2023boundary} induces class-level forgetting by shifting decision boundaries. 
In \cite{shahlow}, \citeauthor{shahlow} targeted the class unlearning problem based on sparse representation learned by Discrete Key-Value Bottleneck \cite{trauble2023discrete}.

Despite differing in their intervention points (weights, representations, or boundaries), these approaches typically depend on the original task formulation and label supervision to define the forgetting objective and drive the update. In contrast, we study representation unlearning without using label information.




%% file: Contents/3_problem_definition.tex

\section{Priliminary and Problem Reformulation} \label{priliminary}

\subsection{Background of Manifold Capacity Representation Theory}

Consider $P$ class-labeled manifolds embedded in a $D$ dimensional feature space. Manifold capacity theory asks: \textit{what is the largest load ratio $\frac{P}{D}$ for which, with high probability, a single hyperplane can separate an arbitrary (random) dichotomy of those manifolds~\citep{gardner1988space}?} Recent work shows that there exists a manifold capacity value $C$, such that the probability of finding a separating hyperplane is approximately 1 when $\frac{P}{D} \leq C$, and essentially 0 when $\frac{P}{D} \geq C$ \citep{chung2018classification}. This critical capacity $C$ can be predicted from three geometric quantities: (1) the manifold radius $R_M$, (2) the manifold dimensionality $D_M$, and (3) the correlation of the manifold centroids. 

When the manifold centroid correlation is low, the manifold capacity value can be approximated by $\phi (R_M \sqrt{D_M})$, where $\phi$ is a monotonically decreasing function. \citeauthor{yerxa2023learning}~\citep{yerxa2023learning} further rewrite the capacity as $C \approx \phi(\sum_{i} \sigma_i)$, where $\sigma_i$ are the singular values of a matrix of sampled manifold points (equivalently, the square roots of the covariance-matrix eigenvalues). Maximizing the manifold capacity representation supports high-quality object recognition. We demonstrate the comparison of training a self-supervise model with or without MMCRs \citep{yerxa2023learning} on MNIST in \Cref{representations_w_or_w_o_mmcrs} in \Cref{t_mmcr}. With MMCRs during model training, the representation will make every class's points stay close together (small intra-class distance) while different classes lie far apart (large inter-class distance). The representation with these properties inspires us to investigate an efficient approximate unlearning method within only the manifold representation.

 \subsection{Problem Reformulation} \label{problem_df}
 
 \noindent
 \textbf{Notation and Setup.} We summarize most notations in \Cref{tab:notation_manif_smc} in \Cref{notations_in_ManiF_SMC}.
 Let $\theta_o$ be the original parameters of the model $f_{\theta_o}(\cdot)$ and $\theta_u$ be the new parameters after unlearning. We denote the original representation of $x_i$ by $\texttt{z}_{i,o} = f_{\theta_o}(x_i)$ and the representation after unlearning as $\texttt{z}_{i,u} = f_{\theta_u}(x_i)$. The original training set $S$ is partitioned into the unlearning set $S_u$ and the remaining set $S_r$. 
 
\noindent
\textbf{Neighborhoods and Local Centroid.}
For each erased sample $x_i\in S_u$, let $S_k^i\subseteq S_r$ be its top-$k$ most similar retained samples,
selected using the original representations (e.g., nearest neighbors of $\texttt{z}_{i,o}$ among $\{\texttt{z}_{j,o}:x_j\in S_r\}$).
We define the local centroid of the top-k similar samples of $x_i$ on the unlearned model as 
\begin{equation} \label{centroid_eq}	
	\texttt{c}_{i,u} \;=\; \frac{1}{|S_k^i|}\sum_{x_k^i\in S_k^i} f_{\theta_u}(x_k^i).
\end{equation}
The original unlearned centroid $\texttt{c}_{i,o}$ for $S_k$ can also be calculated by \Cref{centroid_eq} using the original model $f_{\theta_o}(\cdot)$.

Motivated by the observation in \Cref{representation_retrain_observation}, we use a manifold view: after retraining on $S_r$, an erased sample tends to behave similarly to its semantic neighbors in $S_r$. This suggests that effective approximate unlearning should (i) move the erased sample away from its original representation region, while (ii) aligning it with similar retained samples.

 \noindent
 \textbf{Push Away from the Original Representation.}
 Prior work shows that approximate unlearning can be induced by altering erased samples’ labels or representations
 \citep{fan2024salun,chundawat2023can,guo2019certified}.
 In representation space, this can be expressed by encouraging $\texttt{z}_{i,u}$ to move away from its original representation:
 \begin{equation} \label{representation_unl_target}
 	\textbf{Push term:}\quad
 	\max_{\theta_u}\;\sum_{x_i\in S_u}\big\| f_{\theta_u}(x_i)-\texttt{z}_{i,o}\big\|^2.
 \end{equation}
 However, as we mentioned before, reversing the label or blindly pushing the representation can conflict with the learning objective and may deviate from retraining behavior.

\noindent
\textbf{Pull Toward Similar Retained Samples.} 
To align with the retraining observation, we encourage each erased sample to move toward its semantic neighbors in $S_r$. Using the same neighbor set $S_k^i$, we pull $\texttt{z}_{i,u}$ toward the centroid of those retained neighbors:
 \begin{equation} \label{representation_unl_toward}
 	\textbf{Pull term:}\quad
 	\min_{\theta_u}\;\sum_{x_i\in S_u}\big\| f_{\theta_u}(x_i)-\texttt{c}_{i,u}\big\|^2.
 \end{equation}
 
\noindent
\textbf{Representation-based Approximate Unlearning.}
Together, the push and pull terms (\Cref{representation_unl_target,representation_unl_toward}) define a representation-space approximation to standard retraining: the erased samples are displaced from their original local region while being absorbed into the manifold supported by similar retained data. This reformulation operates purely in representation space and therefore reduces reliance on label perturbations or task-specific gradient manipulation.

%% file: Contents/4_approach.tex
\begin{figure*}[t]
	\centering
	\includegraphics[width=0.8149\linewidth]{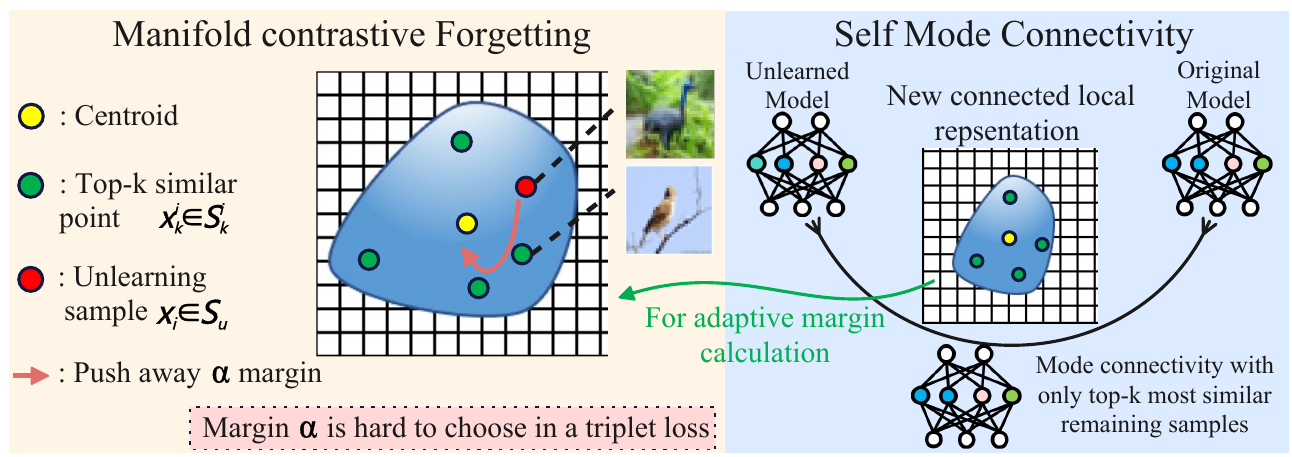}
	\vspace{-2mm}
	\caption{ManiF-SMC for manifold forgetting. ManiF forgets a target sample by pushing itself away from the original representation toward the centroid of its top-$k$ nearest retained samples with margin $\alpha$. To obtain a suitable $\alpha$ for ManiF optimization, we propose self mode connectivity, which links the unlearned model and the original model using only the top-k similar remaining samples, reconstructing a new local representation without the erased data. We use the local representation reconstructed by the new connected model to guide the adaptive margin calculation. 
		\vspace{-2mm}
	}
	\label{fig_figure2}
\end{figure*}

\section{The Manifold Forgetting Guided by Self Mode Connectivity}
\label{sec_triplet_unlearning}



To address the representation-based unlearning objective from a manifold perspective, we propose ManiF-SMC (\Cref{fig_figure2}), which consists of two components:
\begin{itemize}[itemsep=0pt, parsep=0pt, leftmargin=*]
	\item \textbf{Manifold contrastive forgetting (ManiF):} a representation-space unlearning method.
	\item \textbf{Self mode connectivity (SMC):} a module that adaptively calculates the contrastive margin to guide the ManiF optimization.
\end{itemize}

\subsection{Manifold Contrastive Forgetting (ManiF)}

Recall from the previous introduction that the representation-based unlearning goal is maximizing $\sum_{i \in S_u} \| f_{\theta_u}(x_i) - \texttt{z}_{i,o} \|^2,$ and minimizing $\sum_{i \in S_u} \| f_{\theta_u}(x_i) - \texttt{c}_{i,u} \|^2$. The relational push-pull objective is naturally integrated into a triplet or margin-ranking formulation~\citep{schroff2015facenet, weinberger2009distance}. 
We can formulate a margin-based triplet loss for representation-based approximate unlearning as
\begin{equation}
	\label{eq:triplet_loss}
	\begin{aligned}
	\mathcal{L}_{\text{triplet}}(\theta_u) \;=\; 
		\sum_{x_i \in S_u} 
		\Bigl[
		\,\mathrm{dist}\bigl(f_{\theta_u}(x_i),\, \texttt{c}_{i,u}\bigr) 
		\;-\;
		\mathrm{dist}\bigl(f_{\theta_u}(x_i),\, \texttt{z}_{i,o}\bigr)
		\;+\;  \alpha  
		\Bigr]_+,
	\end{aligned}
\end{equation}
where $[\, \cdot \,]_+ = \max\{\, \cdot,\,0\}$, and $\alpha>0$ is a margin. Here, $\mathrm{dist}(\cdot,\cdot)$ is commonly the Euclidean distance, and other metrics such as cosine similarity, Kullback-Leibler Divergence, and mutual information could be used too. Minimizing $\mathcal{L}_{\text{triplet}}$ enforces: 
\begin{equation}
	\label{eq:triplet_margin}
	\mathrm{dist}(f_{\theta_u}(x_i),\, \texttt{c}_{i,u}) \;+\;\alpha \;\;\le\;\; \mathrm{dist}(f_{\theta_u}(x_i),\, \texttt{z}_{i,o}).
\end{equation}
Hence, the unlearned new representation for $x_i$ is at least $\alpha$ closer to the top-k remaining similar samples' centroid than to the original representation. Based on the model previously learned compact manifold representation, we have the following unlearning distance guarantee.

 \noindent
\textbf{Unlearning Lower Bound of ManiF.}
The triplet loss in \Cref{eq:triplet_loss} is a sum of hinge terms and is always non-negative. Hence, $\mathcal{L}_{\text{triplet}}(\theta_u)=0$ if and only if each hinge term is zero, i.e., for all $x_i\in S_u$,
\begin{equation}
	\mathrm{dist}(f_{\theta_u}(x_i),\, \texttt{c}_{i,u}) \;+\;\alpha \;\;\le\;\; \mathrm{dist}(f_{\theta_u}(x_i),\,\texttt{z}_{i,o}).
\end{equation}
Thus, $\alpha$ provides a margin-based lower bound: the unlearned representation of $x_i$ is at least $\alpha$ closer to the retained-neighbor centroid than to its original representation (under $\mathrm{dist}$).

Ideally, if we have a retrained model on $S_r$, we can use it to estimate the post-retraining neighbor centroid $\texttt{c}_{i,u}$ and choose a suitable margin $\alpha$. However, we don't have the retrained model. \textit{Can we have a fast replacement?} We propose to use the local mode connectivity \citep{garipov2018loss,zhao2020bridging} to fastly reconstruct a local manifold representation for the top-k most similar samples $S_{k}^i$ in the remaining dataset.


\subsection{Adaptive Margin for ManiF Guided by Self Mode Connectivity} \label{SCM_section}

\textbf{Intuition.}
In our problem, we need a surrogate model to reconstruct a local representation that is largely unaffected by the unlearned samples, so that we can use it to guide the post-unlearning neighbor centroid and margin calculation. Mode connectivity shows that well-trained solutions can be linked by a low-loss path, enabling efficient sampling of high-performing models without full retraining \citep{garipov2018loss}. Prior work demonstrates that when two models are poisoned, learning a path between them using only a limited amount of benign data can produce intermediate models that mitigate the adversarial behavior while preserving clean accuracy \citep{zhao2020bridging}. 

Motivated by this, we learn a local connecting path using only the retained neighborhood data (e.g., $S_k=\cup_{x_i\in S_u} S_k^i$) and use a sampled model on this path to compute the centroid and an adaptive margin.

\noindent
\textbf{Self Mode Connectivity via a Quadratic Bézier Path.}
Given endpoints $\theta_u$ and $\theta_o$, we parameterize a quadratic Bézier curve
\begin{equation}\label{mc_two}
	\phi_w(t) = (1-t)^2\theta_u + 2t(1-t)w + t^2\theta_o,\quad 0\le t\le 1,
\end{equation}
where $w$ is the learnable control point.
We learn $w$ by minimizing the retained loss along the path (computed on $S_k$), e.g.,
\begin{equation}\label{eq:w_opt}
	\min_w\ \mathbb{E}_{t\sim \mathcal{U}[0,1]}\big[\mathcal{L}_{S_k}(\phi_w(t))\big],
\end{equation}
where $t\sim \mathcal{U}[0,1]$ means $t$ is sampled from the uniform distribution on the interval $[0,1]$.
We then select a surrogate model on the path, $\tilde{\theta}=\phi_w(t^\star)$ (we use $t^\star=0.5$ in practice),
to estimate retained-neighborhood representations.

\noindent
\textbf{Centroid and Adaptive Margin.}
By using $\tilde{\theta}$ to replace the $\theta_{u}$ in \Cref{centroid_eq}, we compute the neighbor centroid $\texttt{c}_{i,\tilde{\theta}}$: 
\begin{equation}
	\texttt{c}_{i,\tilde{\theta}} = \frac{1}{|S_k^i|}\sum_{x_j\in S_k^i} f_{\tilde{\theta}}(x_j).
\end{equation}
Moreover, based on $\tilde{\theta}$, we can set a sample-wise adaptive margin to replace the original fixed $\alpha$ as
\begin{equation}\label{adaptive_margin}
	\alpha_{\tilde{\theta} }(x_i , \texttt{c}_{i,\tilde{\theta}}, \texttt{z}_{i,o})
	=
	\Big[
	\mathrm{dist}\big(f_{\tilde{\theta}}(x_i),\,\texttt{z}_{i,o}\big)
	-
	\mathrm{dist}\big(f_{\tilde{\theta}}(x_i),\,\texttt{c}_{i,\tilde{\theta}}\big)
	\Big]_+,
\end{equation}
and update ManiF as
\begin{equation}\label{eq_triplet_loss_with_function}
		\begin{aligned}
	\mathcal{L}_{\text{triplet}}(\theta_u)
	=
	& \sum_{x_i\in S_u}
	\Big[
	\mathrm{dist}\big(f_{\theta_u}(x_i),\,\texttt{c}_{i,\tilde{\theta}}\big) \\
	& -
	\mathrm{dist}\big(f_{\theta_u}(x_i),\,\texttt{z}_{i,o}\big)
	+ \alpha_{\tilde{\theta}} ( x_i, \texttt{c}_{i,\tilde{\theta}}, \texttt{z}_{i,o})
	\Big]_+.
		\end{aligned}
\end{equation}
We call the resulting method ManiF-SMC and present the whole ManiF-SMC pseudocode in \Cref{algorithm_appendix}.

\noindent
\textbf{Enhancement of Self-Mode-Connectivity (SMC).}
ManiF-SMC uses SMC to obtain a fast surrogate model $\tilde{\theta}$ (sampled on a low-loss Bézier path)
that approximates the retained-data geometry without retraining. This surrogate is used for both (i) estimating the retained-neighbor centroid $\texttt{c}_{i,\tilde{\theta}}$ and (ii) setting a sample-wise adaptive margin $\alpha_{\tilde{\theta}} ( x_i, \texttt{c}_{i,\tilde{\theta}}, \texttt{z}_{i,o})$.
\begin{proposition}[Logit drift along the SMC path]
	\label{prop:smc_logit2}
	Assume $g(\theta,x)$ is $L_x$-Lipschitz in $\theta$. Let $\tilde{\theta}=\phi_w(t^\star)$ on \Cref{mc_two}. Then
	\begin{equation}
	\bigl|g(\tilde{\theta},x)-g(\theta_o,x)\bigr|
	\le L_x\Big((1-t^\star)^2\|\theta_u-\theta_o\|_2 + 2t^\star(1-t^\star)\|w-\theta_o\|_2\Big).
	\end{equation}
\end{proposition}
This smoothness implies that retained-sample predictions (and hence retained-neighbor representations) vary continuously along the path. Therefore, using $\tilde{\theta}=\phi_w(t^\star)$ yields a stable estimate of the retained-neighborhood centroid $\texttt{c}_{i,\tilde{\theta}}$, and the adaptive margin computed under $\tilde{\theta}$ reflects the local retained geometry more faithfully than a fixed global margin.

%% file: Contents/5_experiments.tex

\section{Experiments} \label{exp}

In this section, we conduct experiments to answer the following research questions (RQs) and evaluate ManiF-SMC:
\begin{itemize}[itemsep=0pt, parsep=0pt, leftmargin=*]
	\item \emph{\textbf{RQ1}}: How does ManiF-SMC perform in terms of unlearning effectiveness and efficiency compared with state-of-the-art approximate unlearning methods? (See \Cref{effectiveness_eval})
	\item \emph{\textbf{RQ2}}: What are the contributions of key components (e.g., MMCR and SMC-based adaptive margin), and how do different parameters and metrics influence the ManiF-SMC? (See \Cref{MMCR_eval,SMC_eval})
	\item \emph{\textbf{RQ3}}: How well does ManiF-SMC generalize to more challenging settings, such as unlearning for generative models and label-limited scenarios where task labels are unavailable? (See \Cref{Generative_eval,access_limitation})
\end{itemize}

\subsection{Experimental Setting}

\noindent
\textbf{Datasets.}
We have conducted experiments on four widely adopted public datasets: MNIST \citep{deng2012mnist}, CIFAR10 \citep{krizhevsky2009learning}, CelebA \citep{liu2018large}, and Tiny-ImageNet \citep{le2015tiny}, offering a range of objective categories with varying levels of learning complexity. We present detailed statistics of all datasets and how do we use them in \Cref{datasets_appendix}.



\noindent
\textbf{Models.}
We select three model architectures of different sizes in our experiments: a 5-layer multi-layer perceptron~(MLP) connected by ReLU, a 7-layer convolutional neural network (CNN), and ResNet-18. We use ResNet-18 as an encoder to learn the manifold representation and connect with a MLP for task models on MNIST and a CCN for task models on CIFAR10, CelebA, and Tiny-ImageNet. We set the dimensionality of learned manifold representation $D_M = 5$ on MNIST, and $D_M = 64$ on CIFAR10 and CelebA, and $D_M = 256$ on Tiny-ImageNet. During training, we set the minibatch size to $16$ on MNIST, CIFAR10, and CelebA, and the minibatch size to $200$ on Tiny-Imagenet. All the experiments are conducted on NVIDIA Quadro RTX 6000 GPUs.


\noindent
\textbf{Metric.}
Existing studies have assessed machine unlearning performance from different aspects~\citep{graves2021amnesiac,golatkar2020eternal}. By carefully reviewing the prior arts, we focus on the empirical metrics including Membership inference attack \textbf{(MIA)}, Remaining accuracy \textbf{(RA)}, Testing accuracy \textbf{(TA)},  Remaining mean-squared error \textbf{(R-MSE)}, Testing mean-squared error \textbf{(T-MSE)}, and Running Time \textbf{(RT)}. We demonstrate the detailed introduction of these metrics in \Cref{metric_detail}.

\noindent
\textbf{Compared Unlearning Benchmarks.} We compare our method with four mainstream unlearning algorithms: \textbf{Retraining}, \textbf{GA}~\citep{graves2021amnesiac,thudi2022unrolling}, \textbf{VBU}~\citep{nguyen2020variational}, \textbf{RFU}~\citep{wang2023machine} and \textbf{SalUn}~\citep{fan2024salun}. The corresponding introduction of each method is provided \Cref{revisiting_mu}.

 \begin{figure}[t]
 	\centering
 	\vspace{-2mm}
 	\hspace{-2mm}
 	\subfloat{	\label{fig:garandomforgetting10pcifar1040}
 		\includegraphics[scale=0.242]{Contents/figures/GA_retrain_TSNE/GA_random_forgetting10p_CIFAR10}
 	}			
 	\hspace{-4mm}
 	\subfloat{  	\label{fig:vburandomforgetting10pcifar10}
 		\includegraphics[scale=0.242]{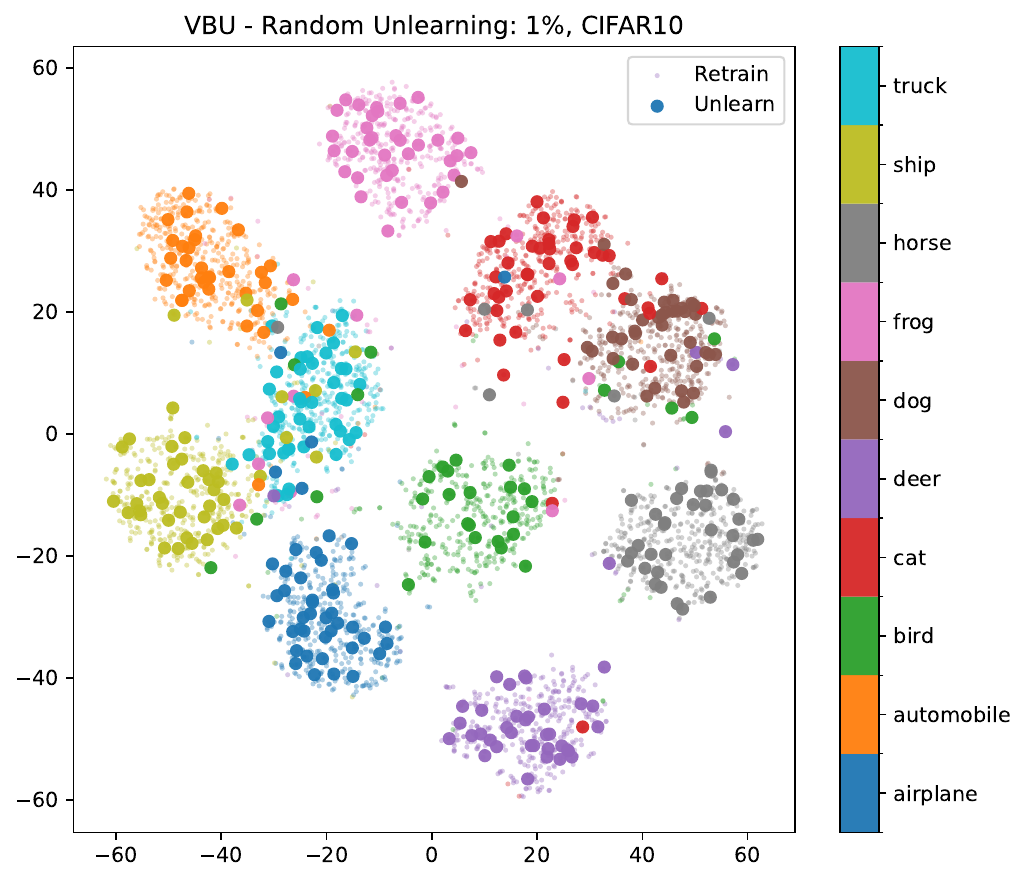}
 	}	
 	\vspace{-2mm} \\
 	\hspace{-2mm}
 	\subfloat{ 		\label{fig:salunrandomforgetting10pcifar10}
 		\includegraphics[scale=0.242]{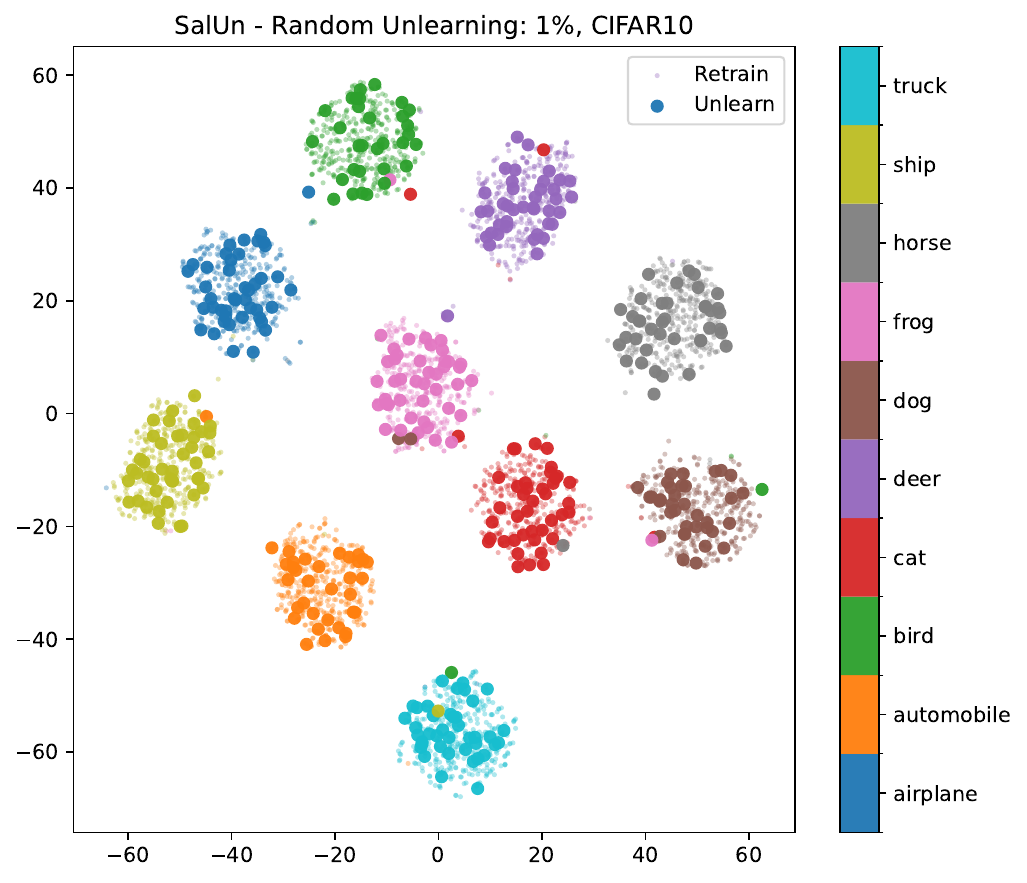}
 	}
 	\hspace{-4mm}
 	\subfloat{ 	  	\label{fig:manifurandomforgetting10pcifar10}
 		\includegraphics[scale=0.242]{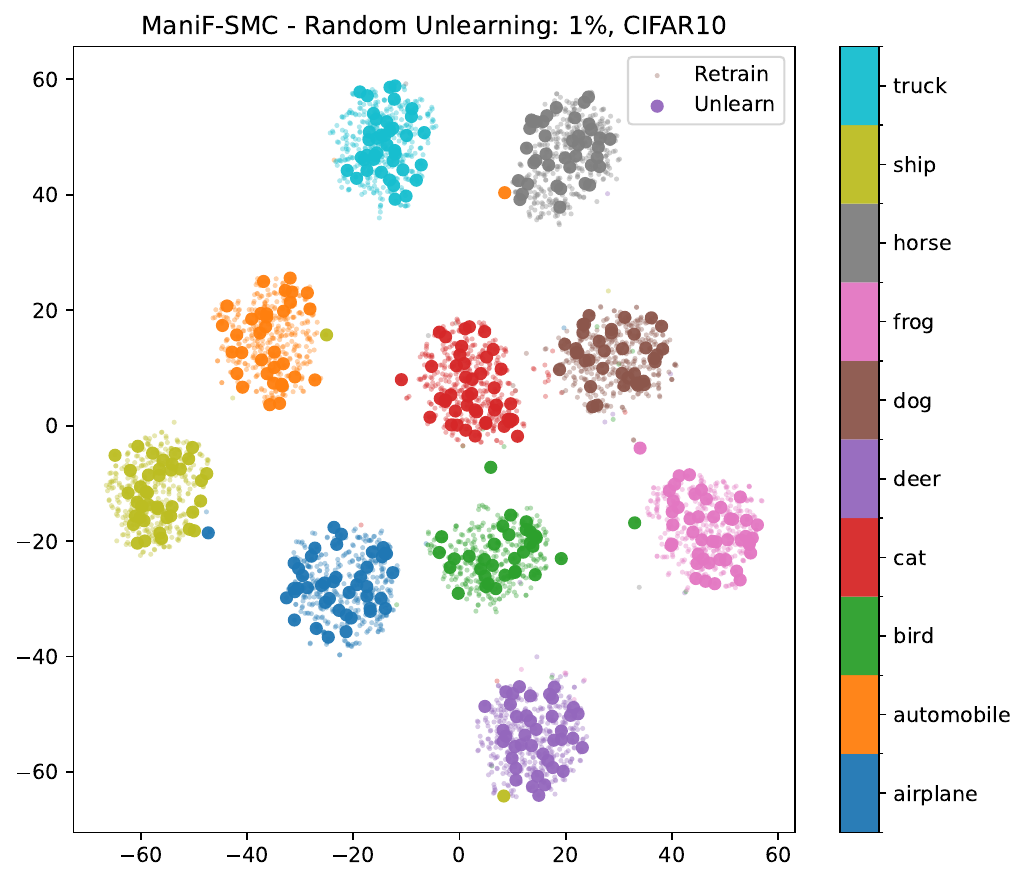}
 	}
 	\vspace{-2mm}
 	\caption{Representation Space of GA, VBU, SalUn and ManiF-SMC on CIFAR10 after unlearning 1\% (500) randomly selected samples. Small points denote retained samples from different clusters. Larger points denote forgotten samples, assigned to the retained cluster with the highest semantic similarity. } 
 	\label{cifar10_representation_space} 
 	\vspace{-2mm}
 \end{figure}

 \begin{figure*}[t]
 	\centering
 	\vspace{-2mm}
 	\hspace{-2mm}
 	\subfloat{	\label{fig:mnistmodelmiauss}    \rotatebox{90}{ \hspace{9mm}	\scriptsize{On MNIST} }
 		\includegraphics[scale=0.23]{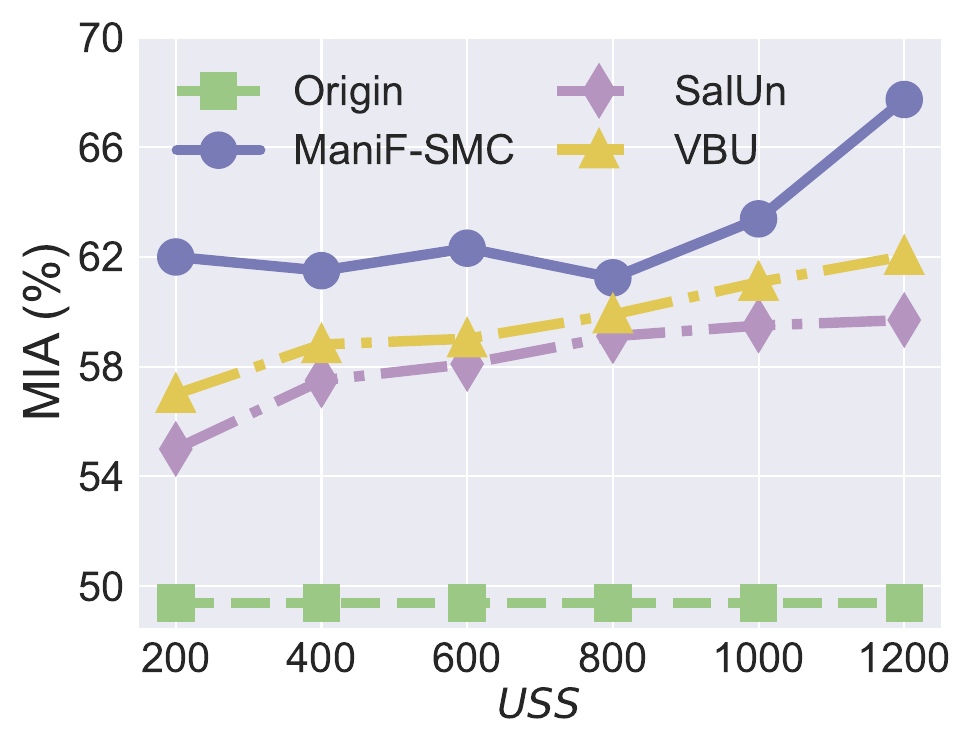}
 	}			
 	\hspace{-2mm}
 	\subfloat{ 		\label{fig:mnistmodelrauss}
 		\includegraphics[scale=0.23]{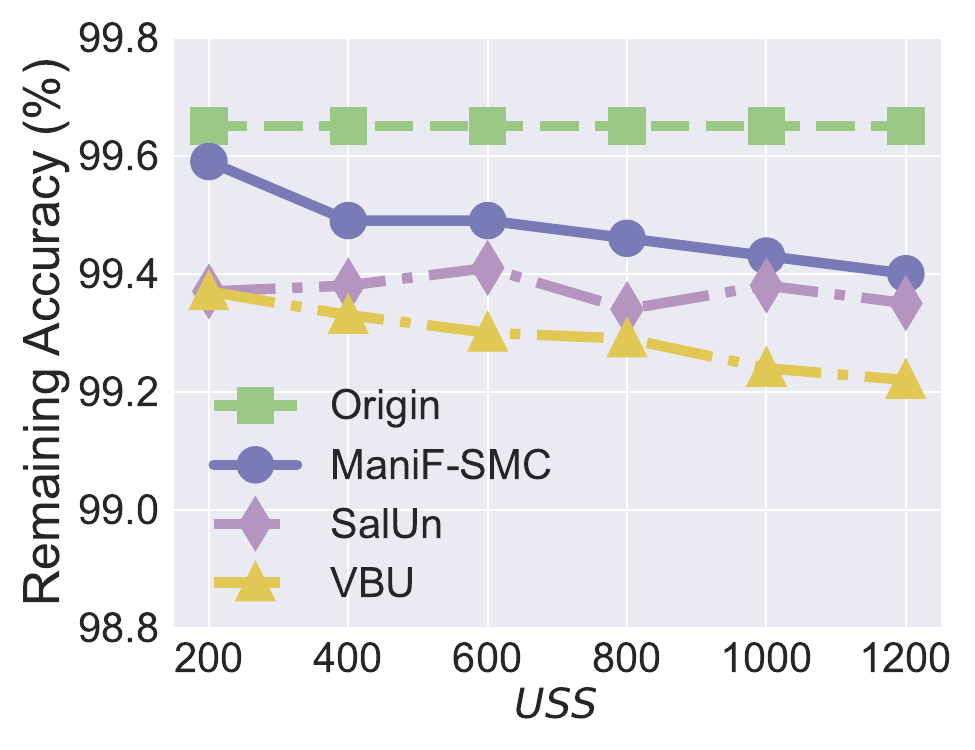}
 	}
 	\hspace{-2mm}
 	\subfloat{  		\label{fig:mnistmodeltauss}
 		\includegraphics[scale=0.23]{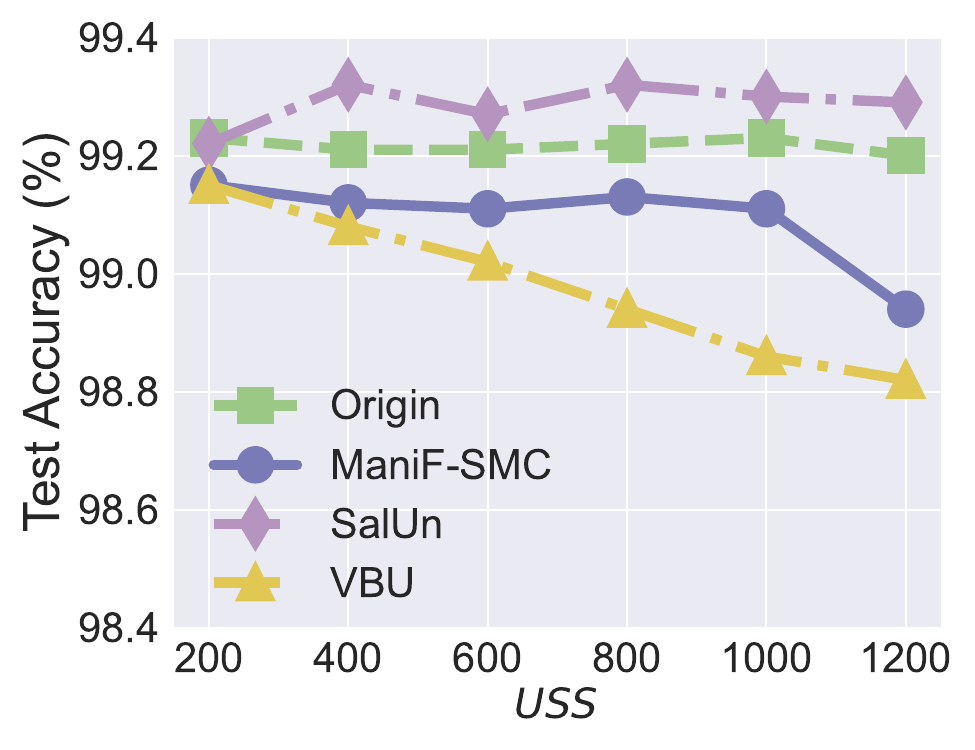}
 	}	
 	\hspace{-2mm}
 	\subfloat{		\label{fig_mnistrunningtimecsasbar}
 		\includegraphics[scale=0.201]{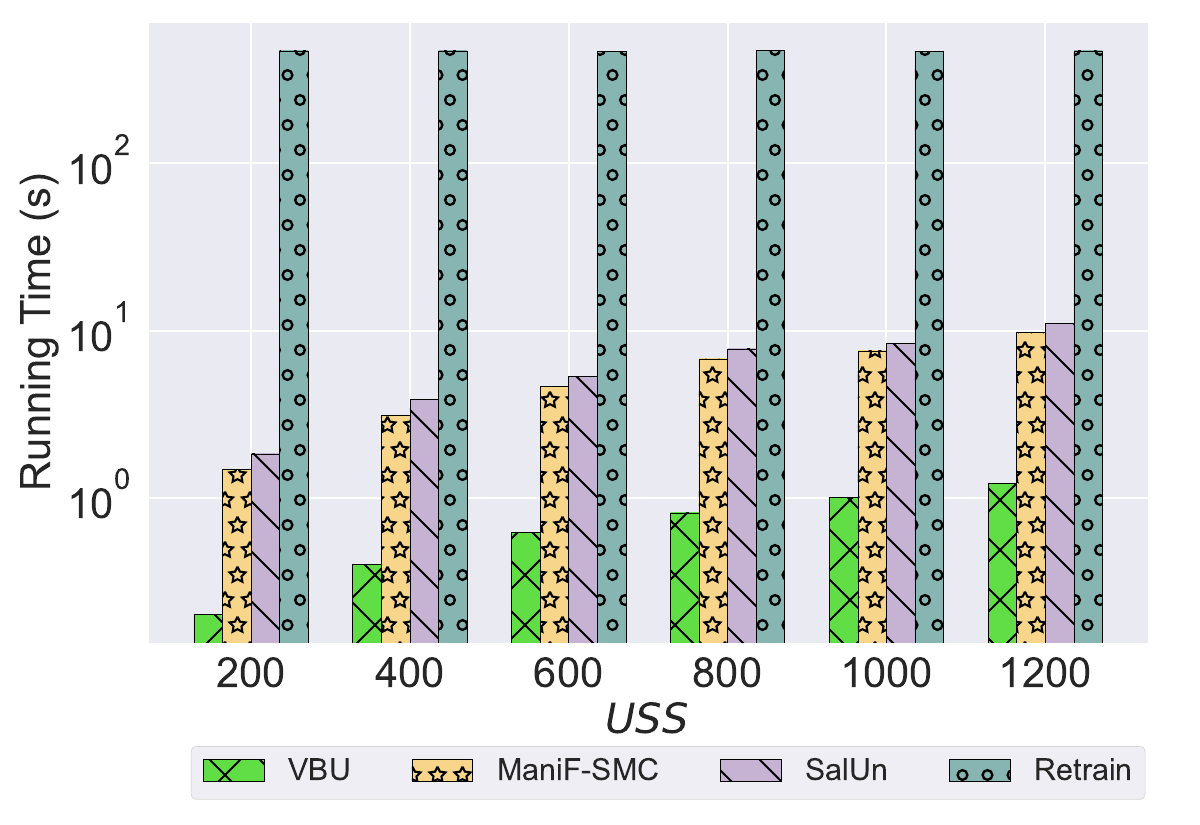}
 	}			
 	\vspace{-5mm} \\
 	\hspace{-2mm}
 	\subfloat{	\label{fig:cifar10modelmiauss}     \rotatebox{90}{ \hspace{9mm}	\scriptsize{On CIFAR10} }
 		\includegraphics[scale=0.23]{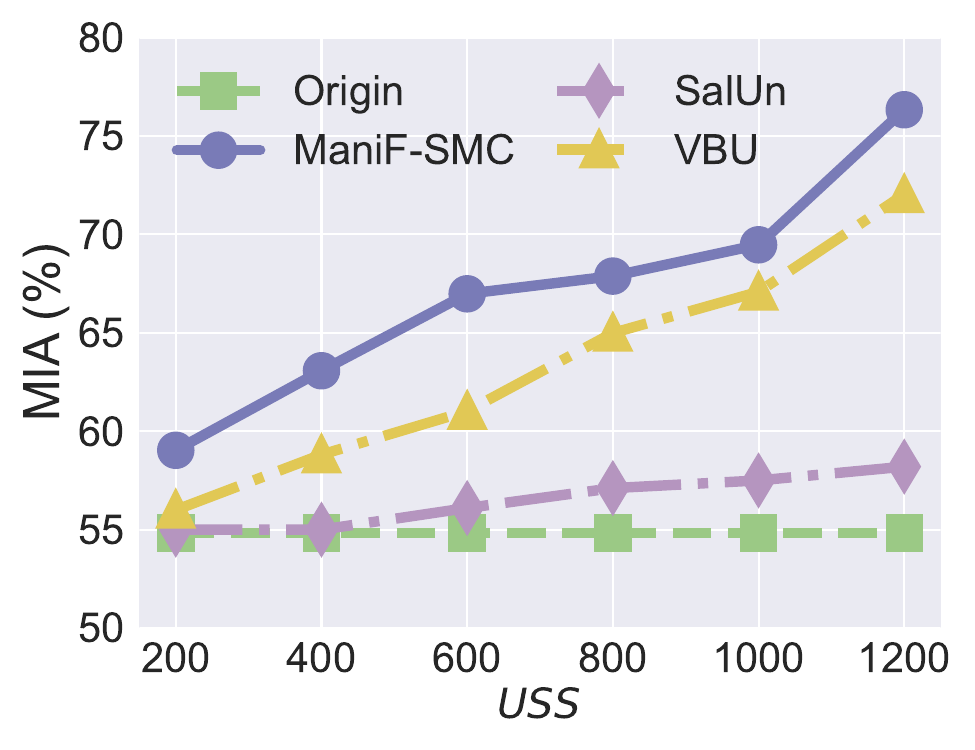}
 	}	
 	\hspace{-2mm}
 	\subfloat{ 		\label{fig:cifar10modelrauss}
 		\includegraphics[scale=0.23]{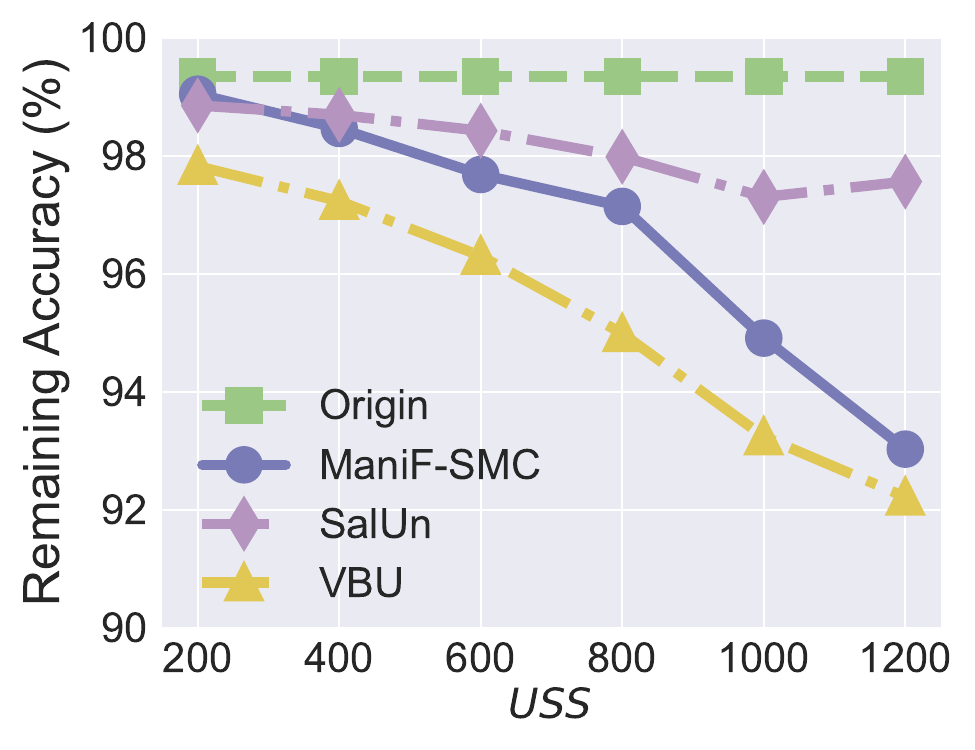}
 	}		
 	\hspace{-2mm}
 	\subfloat{  		\label{fig:cifar10modeltauss}
 		\includegraphics[scale=0.23]{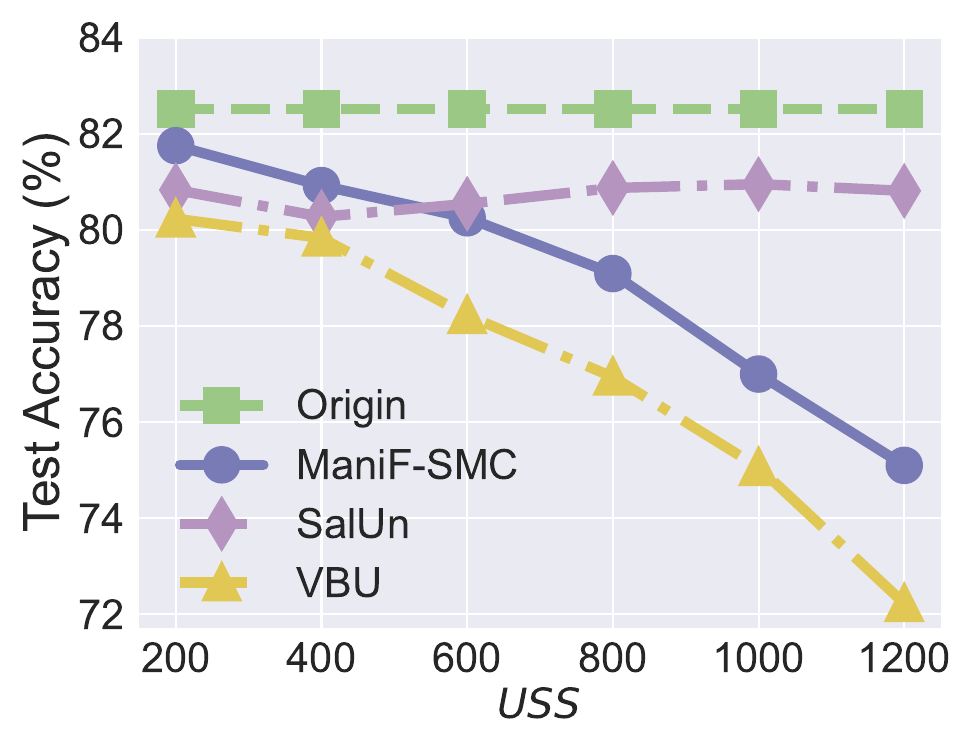}
 	}
 	\hspace{-2mm}
 	\subfloat{ 		\label{fig_cifar10runningtimecsasbar}
 		\includegraphics[scale=0.201]{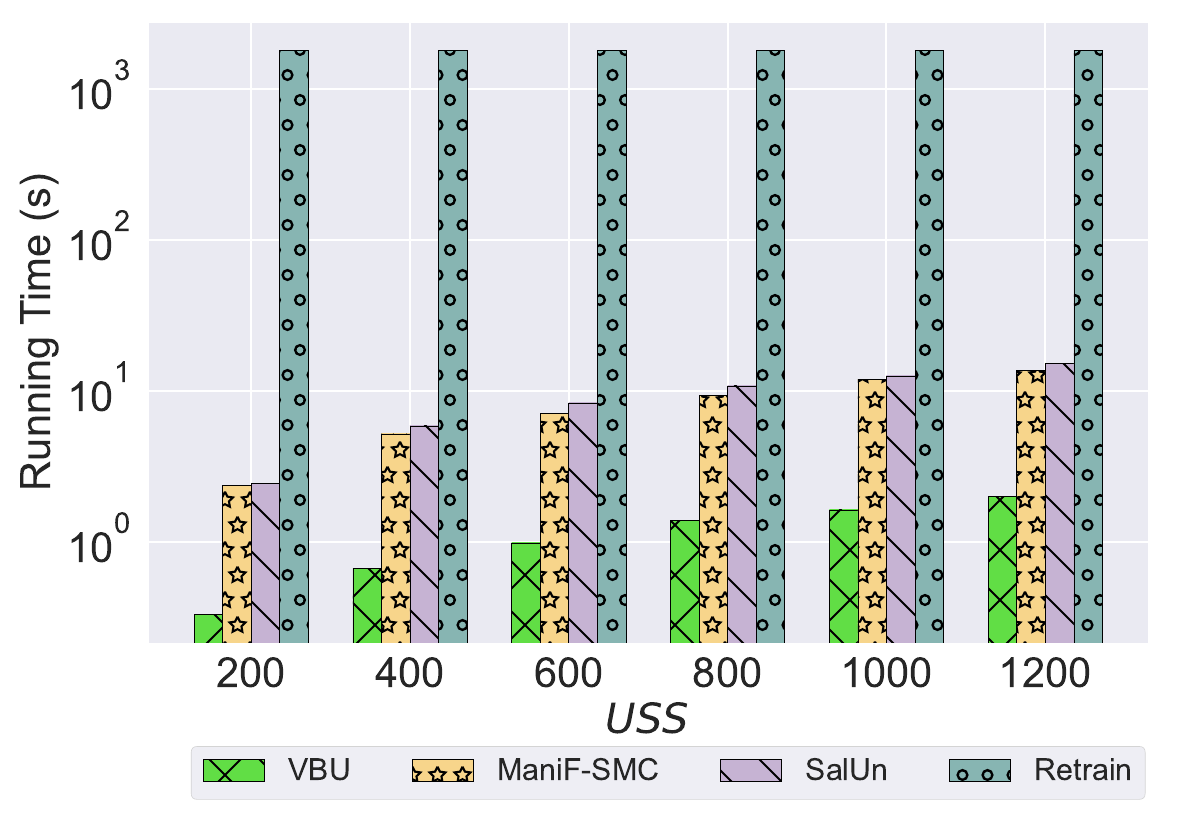}
 	}
 	\vspace{-5mm} \\
 	\subfloat{  	\label{fig:tinyimagenetmodelmiauss}   \rotatebox{90}{ \hspace{6mm}	\scriptsize{On TinyImageNet} }
 		\includegraphics[scale=0.23]{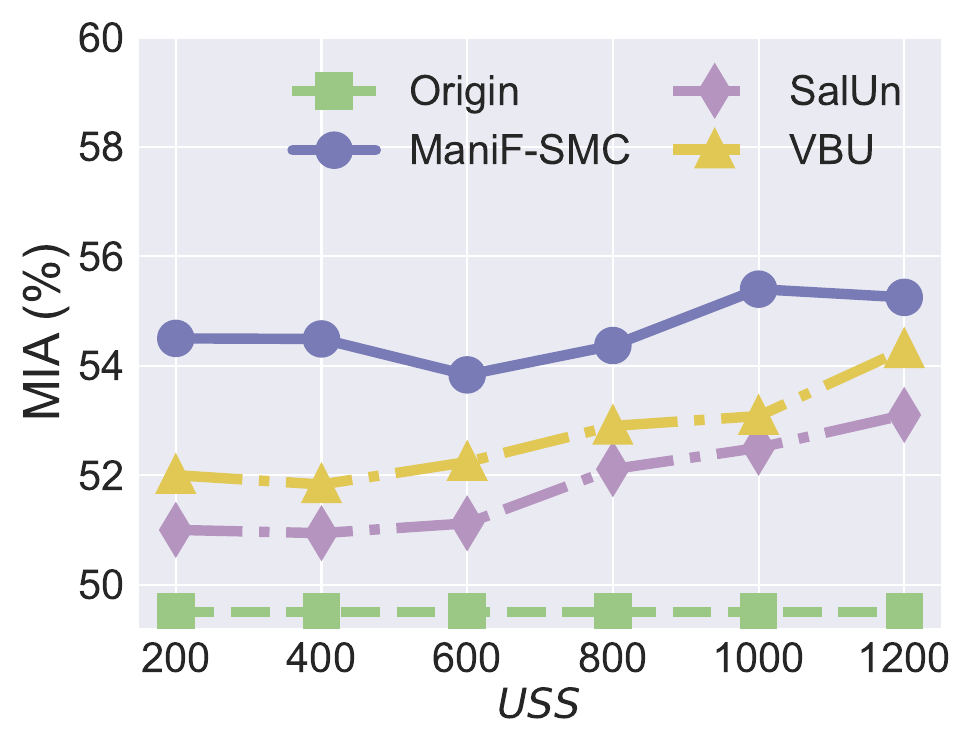}
 	}	
 	\hspace{-2mm}
 	\subfloat{ 	  	\label{fig:tinyimagenetmodelrauss}
 		\includegraphics[scale=0.23]{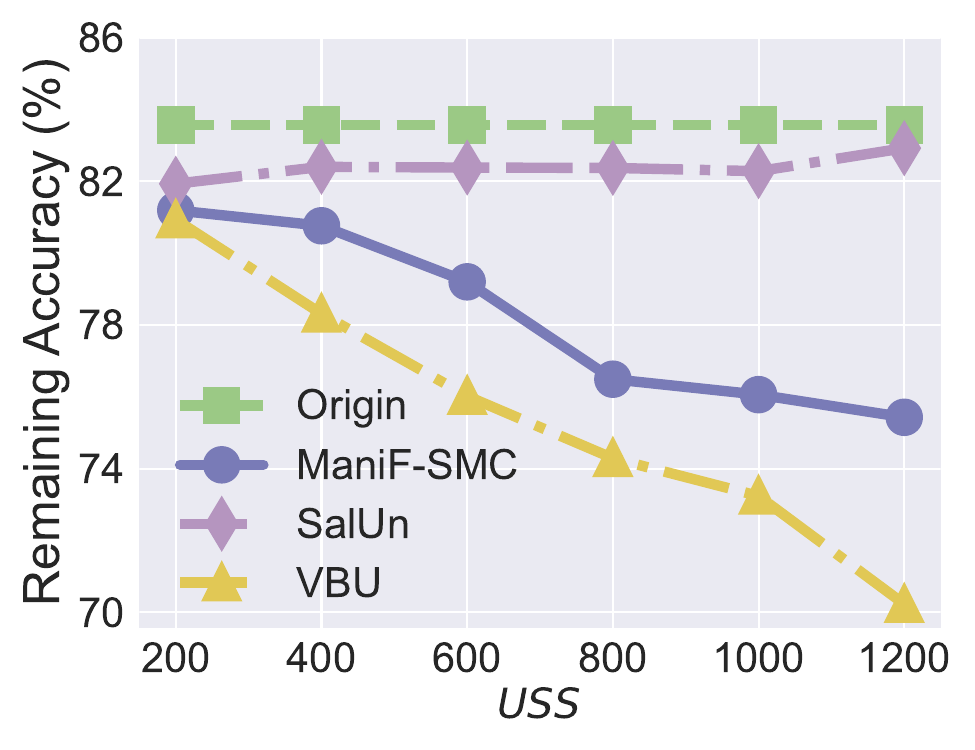}
 	}
 	\hspace{-2mm}
 	\subfloat{  	 	\label{fig:tinyimagenetmodeltauss}
 		\includegraphics[scale=0.23]{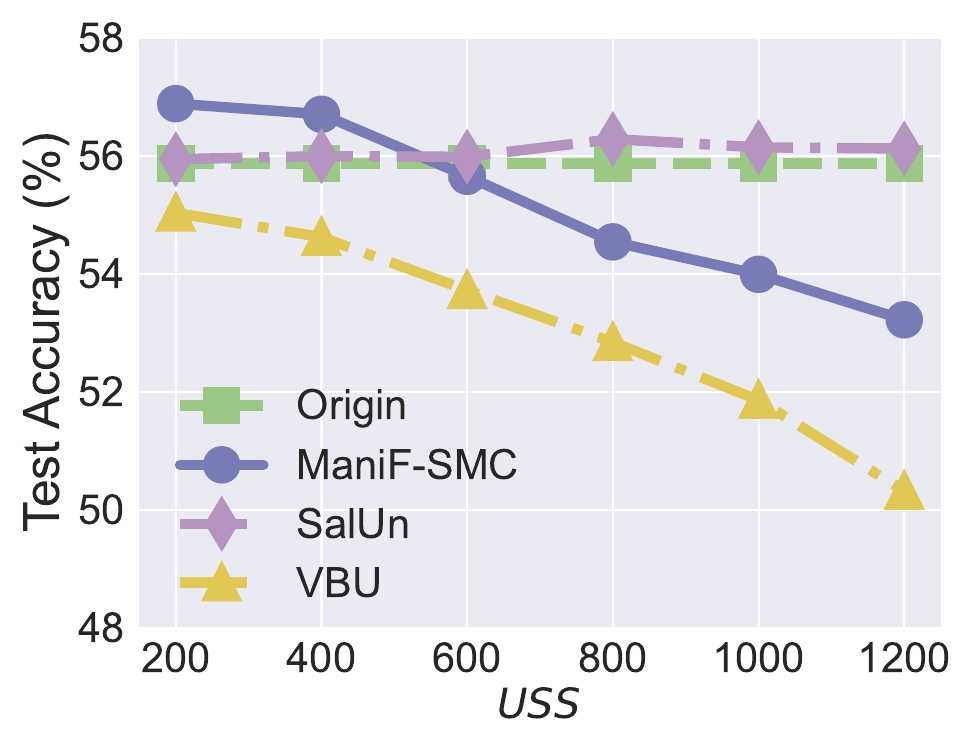}
 	} 	
 	\hspace{-2mm}
 	\subfloat{  	 	\label{fig:tinyimagenet_time}
 		\includegraphics[scale=0.201]{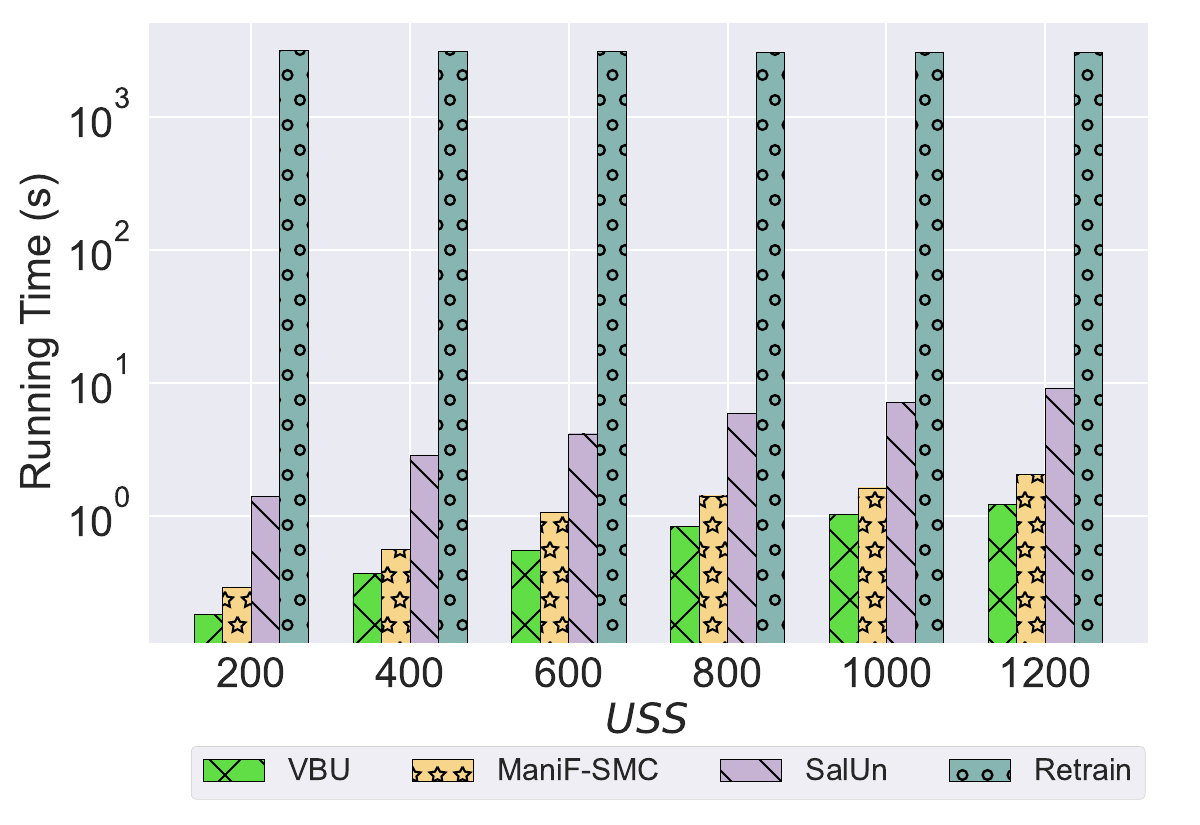}  
 	}
 	\vspace{-2mm}
 	\caption{Overall of unlearning performance on MNIST, CIFAR10, and TinyImageNet. ManiF-SMC is the only method that unlearns without labels; the others require full inputs and labels. SalUn maintains higher utility at larger $\it USS$ by fine-tuning on retained data. ManiF-SMC can be improved similarly (see \Cref{with_finetuning} and \Cref{evaluation_of_mcr_r}). 
 	} 
 	\label{evaluation_of_uss} 
 \end{figure*}

 \subsection{Overview Evaluations} \label{effectiveness_eval}

\noindent
\textbf{Setup.} A common evaluation of unlearning is to test the unlearning effectiveness of different unlearning sample sizes (\textit{USS}). We evaluate how the different unlearning methods perform in various \textit{USS}, settings from 200 to 1200, where 1200 is around $2\%$ training data on MNIST and CIFAR10, already large enough for unlearning according to \citep{chen2021machine,bertram2019five}. 

\noindent
\textbf{Evaluation of Effectiveness.} 
\Cref{cifar10_representation_space} visualizes the unlearned representation space on CIFAR10 when unlearning $1\%$ randomly selected samples. Since GA and VBU implement unlearning with a gradient reversal term, this conflicts with the original learning algorithm, and they increase the intra-class distance. SalUn includes the fine-tuning using the remaining dataset. Hence, SalUn has a more compact class representation than GA and VBU. ManiF-SMC preserves a compact class structure and moves erased samples toward their most semantically similar neighbors in the retained set. Its representation geometry closely resembles SalUn and retraining from scratch (\Cref{representation_retrain_observation}), while requiring no class labels and no fine-tuning on the retained data.

The first to third columns in \Cref{evaluation_of_uss} illustrate how each method's unlearning performance evolves as the unlearning sample size $\it USS$ increases on MNIST, CIFAR10, and Tiny-ImageNet.
Additional results on CelebA are presented in \Cref{evaluation_of_uss_cifar10_celeba} in \Cref{add_exp_eff_uss}. In the first column of plots, we observe that the MIA rate rises with larger \textit{USS} in all methods, indicating that removing more data can increase the unlearning effectiveness, a higher rate that MIA recognizes not in the training dataset. All unlearning methods (ManiF-SMC, VBU, and SalUn) can effectively unlearn the original models. 

The second and third columns track RA on the retrained data and TA on the test data, respectively. As $\it USS$ increases, both ManiF-SMC and VBU show slight drops in RA and TA, which indicates that these two approximate unlearning methods will degrade model utility to some extent. ManiF-SMC tends to preserve accuracy better than VBU on these datasets and only uses the manifold representation. 

SalUn can preserve a good model utility when $\it USS$ increases because it utilizes the remaining data to fine-tune the model during unlearning. We can also add the finetuning of the remaining dataset to mitigate the model utility degradation of ManiF-SMC, where the corresponding results are presented in \Cref{with_finetuning}.

\noindent
\textbf{Impact on Efficiency.} The fourth column in \Cref{evaluation_of_uss} provides a detailed view of the running time, shown on a logarithmic scale. These results reveal that retraining is by far the most time-consuming approach, requiring on the order of more than $10^2$ seconds on MNIST, and $10^3$ to $10^4$ seconds on CIFAR10 and TinyImageNet. In contrast, ManiF-SMC, VBU, and SalUn remain under one second, even for larger \textit{USS} values.

\begin{table*}[t]
	\centering
	\caption{ Ablation study of various machine unlearning methods with (w) and without (w/o) MMCRs. We randomly choose 200 samples from different classes to unlearn. The results with \textcolor{blue}{blue} color show how much the MMCRs improve, and the results with \textcolor{red}{red} color show the negative influence caused by MMCRs. Best and second-best results are highlighted in \textbf{Bold} and \textit{italic}, respectively. ManiF-SMC generally ranks first or second.
		\vspace{-2mm}
	}
	\label{tab_mu_performance}
	\resizebox{\textwidth}{!}{
		\setlength\tabcolsep{2.pt}
		\begin{tabular}{|c|c|cc|cc|cc|cc|}
			\toprule[1pt]
			\toprule[1pt]
			\multirow{2}{*} { \makecell[c]{\textbf{Datasets} } } & \multirow{2}{*} { \makecell[c]{\textbf{Unlearning} \\ \textbf{Methods} } } & \multicolumn{2}{c}{\textbf{MIA($\%$)}} & \multicolumn{2}{c}{\textbf{RA ($\%$)}} & \multicolumn{2}{c}{\textbf{TA ($\%$)}} & \multicolumn{2}{c}{\textbf{RT} (second)}\\
			\cmidrule(lr){3-4}
			\cmidrule(lr){5-6}
			\cmidrule(lr){7-8}
			\cmidrule(lr){9-10}
			& & \textbf{w/o MMCRs} & \textbf{w MMCRs} 
			& \textbf{w/o MMCRs} & \textbf{w MMCRs} 
			& \textbf{w/o MMCRs} & \textbf{w MMCRs} 
			& \textbf{w/o MMCRs} & \textbf{w MMCRs}  \\
			\midrule[0.12em]
			\multirow{5}{*} {  \makecell[c]{\textbf{ \rotatebox{90}{ On MNIST }  } } }
			&\textbf{Retraining} 
			& 56.00   & \textit{63.00} (\textcolor{blue}{$\uparrow$ 7.00})
			& 99.24  & 99.49 (\textcolor{blue}{$\uparrow$ 0.25})
			& 99.03  & \textit{99.23} (\textcolor{blue}{$\uparrow$ 0.20})  
			& 464.9   & 470.4    (\textcolor{red}{$\uparrow$ 5.5})\\
			& \textbf{GA}      
			& \textbf{60.00} & \textbf{64.50} (\textcolor{blue}{$\uparrow$ 4.50})
			& 99.08  & 99.01 (\textcolor{red}{$\downarrow$ 0.07})
			& 98.77  & 98.81 (\textcolor{blue}{$\uparrow$ 0.04})
			& \textit{0.220} & \textit{0.202} (\textcolor{blue}{$\downarrow$ 0.018})   \\
			&\textbf{VBU}      
			& 56.00  & 57.00 (\textcolor{blue}{$\uparrow$ 1.00})
			& 99.30  & 99.37 (\textcolor{blue}{$\uparrow$ 0.07})
			& 99.05  & 99.15 (\textcolor{blue}{$\uparrow$ 0.10}) 
			& \textbf{0.198}  & \textbf{0.201} (\textcolor{red}{$\uparrow$ 0.003})  \\
			&\textbf{RFU}      
			& 51.00  & 53.50 (\textcolor{blue}{$\uparrow$ 2.50})
			& 99.28  & \textit{99.39} (\textcolor{blue}{$\uparrow$ 0.11})
			& \textbf{99.21}  & \textbf{99.28} (\textcolor{blue}{$\uparrow$ 0.07}) 
			& 0.342  & 0.354 (\textcolor{red}{$\uparrow$ 0.012}) \\
			&\textbf{SalUn}      
			& 53.50  & 55.00 (\textcolor{blue}{$\uparrow$ 1.50})
			& \textit{99.31}  & 99.37 (\textcolor{blue}{$\uparrow$ 0.06})
			& \textit{99.17}  & 99.22 (\textcolor{blue}{$\uparrow$ 0.05}) 
			& 1.832  & 1.839 (\textcolor{red}{$\uparrow$ 0.007}) \\
			&\textbf{ManiF-SMC (Our)}      
			& \textit{59.00}  & 62.00 (\textcolor{blue}{$\uparrow$ 3.00})
			& \textbf{99.40}  & \textbf{99.59} (\textcolor{blue}{$\uparrow$ 0.19})
			& 99.02  & 99.15  (\textcolor{blue}{$\uparrow$ 0.13})
			& 1.486  & 1.482 (\textcolor{blue}{$\downarrow$ 0.004}) \\
			\midrule[0.12em]
			\multirow{5}{*} { \makecell[c]{\textbf{ \rotatebox{90}{  On CIFAR10 }  } } }
			&\textbf{Retraining} 
			& \textbf{61.00} & \textbf{61.00}(\textcolor{blue}{$\uparrow$ 0.00})
			& \textbf{98.92} & \textbf{99.28} (\textcolor{blue}{$\uparrow$ 0.36})
			& \textbf{82.24}  & \textbf{82.48} (\textcolor{blue}{$\uparrow$ 0.24}) 
			& 1742.0 & 1783.4 (\textcolor{red}{$\uparrow$ 41.4}) \\
			&\textbf{GA}      
			& 51.00   & 54.00 (\textcolor{blue}{$\uparrow$ 3.00})
			& 97.51  &  97.83 (\textcolor{blue}{$\uparrow$ 0.32})
			& 78.92  & 79.96 (\textcolor{blue}{$\uparrow$ 1.04}) 
			& \textbf{0.285}   & \textbf{0.273} (\textcolor{blue}{$\downarrow$ 0.012}) \\
			&\textbf{VBU}      
			&  54.00   & 56.00 (\textcolor{blue}{$\uparrow$ 2.00})
			& 97.12  & 97.82 (\textcolor{blue}{$\uparrow$ 0.70})
			& 79.37  & 80.23 (\textcolor{blue}{$\uparrow$ 0.86})
			&  \textit{0.324}  &  \textit{0.331 } (\textcolor{red}{$\uparrow$ 0.007})  \\
			&\textbf{RFU}      
			& \textit{55.00}   & 56.00  (\textcolor{blue}{$\uparrow$ 1.00})
			& 97.78  & 98.57 (\textcolor{blue}{$\uparrow$ 0.79})
			& 79.24  & 80.89 (\textcolor{blue}{$\uparrow$ 1.65}) 
			& 0.503   & 0.522  (\textcolor{red}{$\uparrow$ 0.019}) \\
			&\textbf{SalUn}      
			& 54.00   & 55.00  (\textcolor{blue}{$\uparrow$ 1.00})
			& 98.78  & 98.85 (\textcolor{blue}{$\uparrow$ 0.07})
			& 80.42  & 80.83 (\textcolor{blue}{$\uparrow$ 0.41}) 
			& 2.423   & 2.432  (\textcolor{red}{$\uparrow$ 0.009}) \\
			&\textbf{ManiF-SMC (Our)}      
			& \textit{55.00}   & \textit{59.00} (\textcolor{blue}{$\uparrow$ 4.00})
			& \textit{98.86}  & \textit{99.04} (\textcolor{blue}{$\uparrow$ 0.18})
			& \textit{80.58}  & \textit{81.75} (\textcolor{blue}{$\uparrow$ 1.17})
			& 2.450   &  2.372 (\textcolor{blue}{$\downarrow$ 0.078})   \\
			\bottomrule[1pt]
		\end{tabular}
	} 
	
	\begin{tabbing}
		\textbf{MIA}: Membership Inference Attack; \textbf{RA}: Remaining Accuracy; \textbf{TA}: Testing Accuracy; \textbf{RT}: Runing Time; \\ 
	\end{tabbing}
	\vspace{-2mm}
\end{table*}

\subsection{Ablation Study: Can MMCRs Improve Approximate Unlearning Effectiveness?} \label{MMCR_eval}
\noindent
\textbf{Setup.} We study the impact of Maximum Manifold Capacity Representations (MMCRs) \citep{yerxa2023learning} on the performance of various machine unlearning methods. We compared the methods to unlearn the original model trained with (w) and without (w/o) MMCRs. We randomly choose 200 samples from different classes as the unlearning dataset, and the corresponding results on MNIST and CIFAR10 are demonstrated in \Cref{tab_mu_performance}. 

\noindent
\textbf{Comparison between ``w'' and ``w/o'' MMCRs.} 
When comparing the use of ``w'' and ``w/o'' MMCRs, it is evident that nearly all methods experience improved performance in MIA, RA, and TA when MMCRs are included. For example, on the CIFAR10 dataset, all unlearning methods (Retraining, GA, VBU, RFU, SalUn, and ManiF-SMC) show enhancements in all three effectiveness metrics when MMCRs are employed. MMCRs have improvements not only for approximate unlearning methods, it also improve the model utility of the retraining method. The cost is that the MMCRs will increase the running time for the original model training, hence, the retraining method will be influenced by more computation cost when employing MMCRs. However, since most approximate unlearning methods are not related to the original training process, employing MMCRs will not have too much influence on computation for these approximate unlearning methods.

\noindent
\textbf{Comparison between ManiF-SMC and Other Unlearning Methods.}
Existing approximate unlearning methods have different pros and cons. Let us focus on the regime of the ``w'' MMCRs scenario. We observe that ManiF-SMC achieves the best MIA, RA, and TA on CIFAR10, except for retraining, which has a tradeoff with its efficiency (RT). Although RFU and SalUn achieve better effectiveness than ManiF-SMC on MNIST, we should notice that ManiF-SMC is an approximate unlearning method that only focuses on the data and representations, without relying on the label information and the remaining dataset for fine-tuning. Moreover, GA yields the worst RA and TA since it directly inverses the task loss on the unlearning data, which is harmful to the model utility but beneficial for unlearning (with a high MIA).

 \subsection{Ablation Study: Adaptive Marigin by Self Mode Connectivity versus Fixed Margin} \label{SMC_eval}

\noindent
\textbf{Setup.} We conduct the ablation study on MNIST and CIFAR10 to evaluate the effectiveness of the proposed adaptive margin guided by self mode connectivity (ManiF-SMC) and pure ManiF. For pure ManiF, we set a fixed margin value of  0.01. We evaluate the methods in the \textit{USS} from 200 to 1000, and other settings are the same as those introduced above.

\noindent
\textbf{Results.} In \Cref{tab_smc_performance}, the MIA of ManiF-SMC rises by around $1\%$–$5\%$ for every unlearning sample size on both MNIST and CIFAR10, showing that erased samples have a higher possibility of being classified not in the training dataset after unlearning. Meanwhile, ManiF-SMC preserves much better RA and TA than ManiF, especially when the unlearning size is large, such as $\textit{USS}$ is 800 or 1000. With the achievements above, the running time cost is also higher for ManiF-SMC, as it additionally executes the self mode connectivity module to calculate the margin. 

We put additional ablation studies to evaluate the influence of different distance metrics, neighborhood size $k$, and sampling $t$ of SMC in Appendix, in \Cref{distance_metrics_eval,add_top_k_ablation,t_eval}, respectively.

\begin{table}[t]
	\caption{\small Unlearning VAEs with ManiF-SMC. \vspace{-4mm}
	}
	\label{additional_exp_on_black_b}
	\resizebox{\linewidth}{!}{
		\setlength\tabcolsep{2pt}
		\begin{tabular}{c|c|c|c|c|c|c|c}
			\toprule[0.8pt]
			\midrule
			& Metrics & Original & \textit{USS}  = 200  & 400  & 600  & 800 & 1000 \\
			\midrule
			\multirow{3}{*}{\makecell[c]{ \rotatebox{90}{  	\small{On MNIST} }  } }  & \cellcolor{verylightgray}  MIA  & \cellcolor{verylightgray}  49.39\% & \cellcolor{verylightgray}  59.50\%    & \cellcolor{verylightgray}  58.50\% &  \cellcolor{verylightgray}  61.60\%  &  \cellcolor{verylightgray}  64.49\% &  \cellcolor{verylightgray}  65.53\%         \\  
			& R-MSE   &   0.0357 &  0.0358   &  0.0364 &  0.0366 &   0.0370 &  0.0377     \\  
			& \cellcolor{verylightgray}  T-MSE  & \cellcolor{verylightgray}  0.0360  &  \cellcolor{verylightgray}  0.0361     & \cellcolor{verylightgray}  0.0367   & \cellcolor{verylightgray}  0.0369   &  \cellcolor{verylightgray}  0.0377   &  \cellcolor{verylightgray}  0.0378    \\
			& \makecell[c]{ Generated \\ Samples } 
			& \adjustbox{valign=b}{
				\includegraphics[scale=0.08]{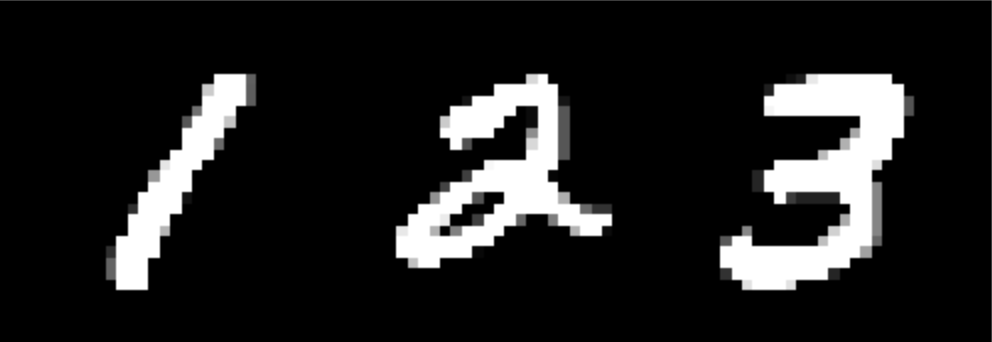} }   
			&   \adjustbox{valign=b}{
				\includegraphics[scale=0.08]{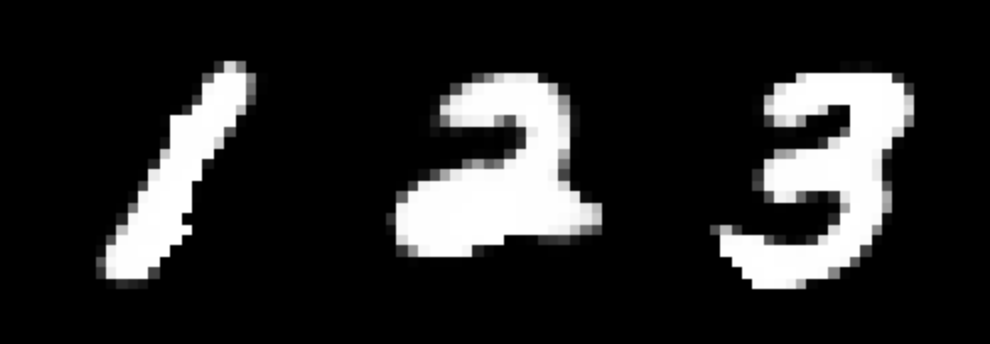} }       
			&  \adjustbox{valign=b}{
				\includegraphics[scale=0.08]{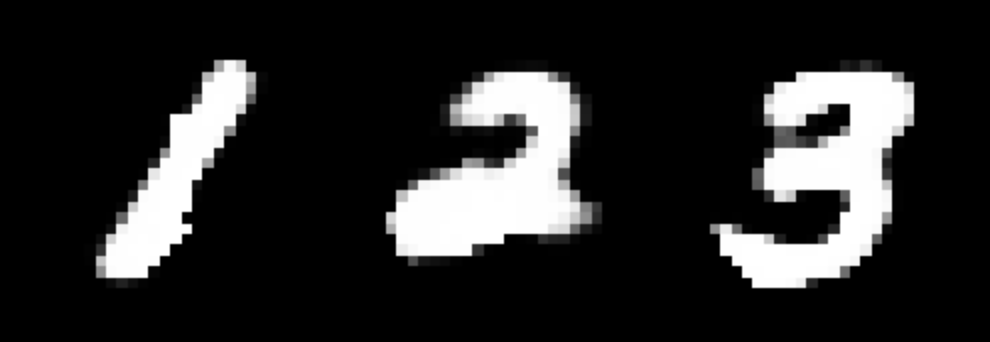} }      
			&  \adjustbox{valign=b}{
				\includegraphics[scale=0.08]{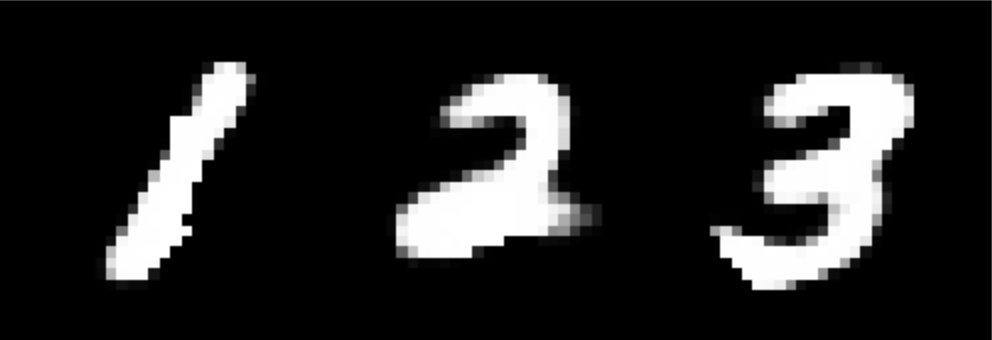} }      
			&   \adjustbox{valign=b}{
				\includegraphics[scale=0.08]{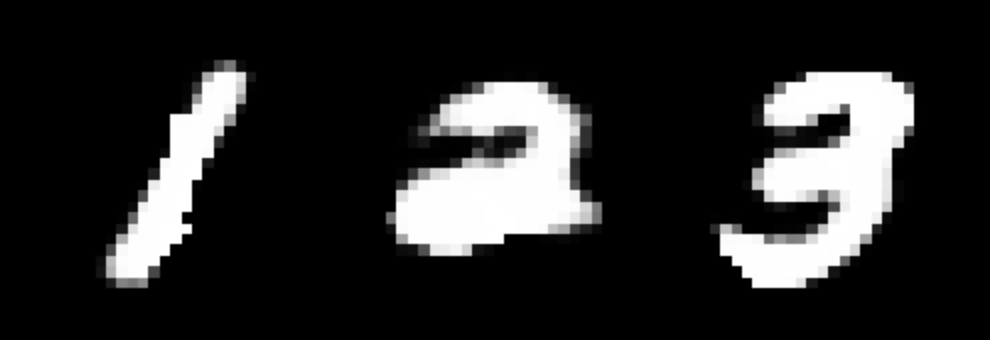} }     
			&   \adjustbox{valign=b}{
				\includegraphics[scale=0.08]{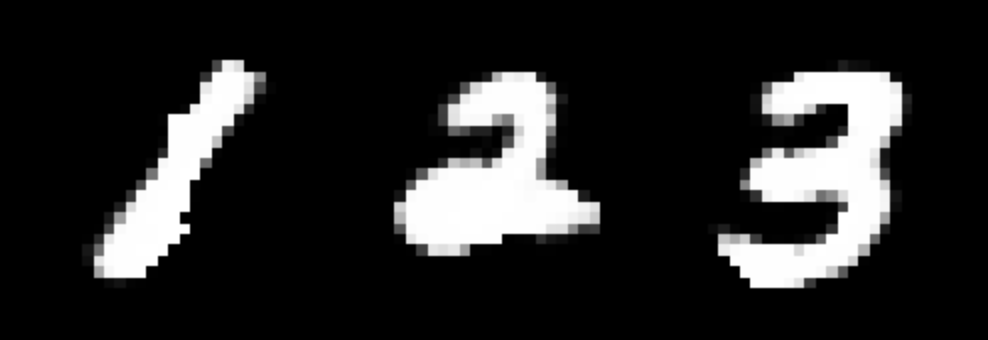} }       \\
			\midrule
			\multirow{3}{*}{\makecell[c]{  \rotatebox{90}{  \hspace{4mm} \small{On CIFAR10} }   } }  & \cellcolor{verylightgray}MIA   & \cellcolor{verylightgray} 53.83\% & \cellcolor{verylightgray} 54.00\%      & \cellcolor{verylightgray} 58.52\%   &  \cellcolor{verylightgray} 57.59\%  & \cellcolor{verylightgray} 61.15\% &  \cellcolor{verylightgray}   66.99\%    \\  
			& R-MSE  & 0.00192   & 0.00192   & 0.00194  &    0.00197   &   0.00213 &   0.00223     \\  
			& \cellcolor{verylightgray} T-MSE  & \cellcolor{verylightgray} 0.00192   & \cellcolor{verylightgray} 0.00192       & \cellcolor{verylightgray} 0.00195 &  \cellcolor{verylightgray} 0.00199  &   \cellcolor{verylightgray} 0.00216  &    \cellcolor{verylightgray} 0.00224    \\
			& \makecell[c]{ Generated \\ Samples } 
			& \adjustbox{valign=b}{
				\includegraphics[scale=0.08]{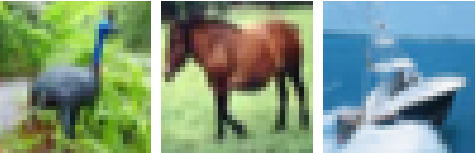} }   
			&   \adjustbox{valign=b}{
				\includegraphics[scale=0.08]{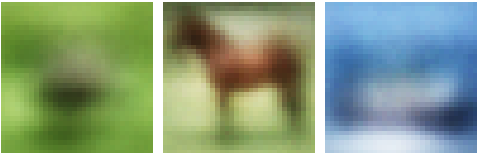} }       
			&  \adjustbox{valign=b}{
				\includegraphics[scale=0.08]{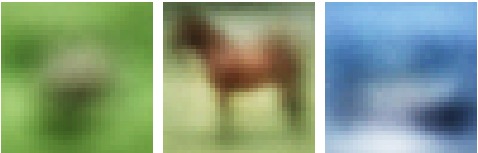} }      
			&  \adjustbox{valign=b}{
				\includegraphics[scale=0.08]{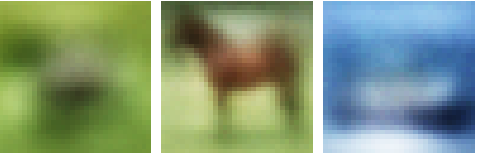} }      
			&   \adjustbox{valign=b}{
				\includegraphics[scale=0.08]{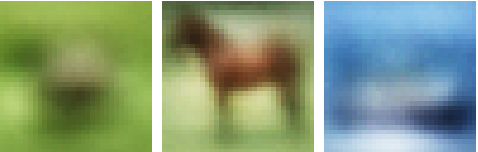} }     
			&   \adjustbox{valign=b}{
				\includegraphics[scale=0.08]{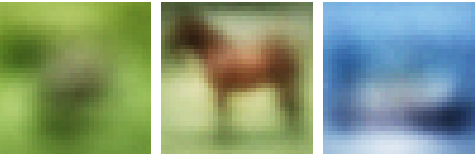} }       \\
			\bottomrule[0.8pt]
	\end{tabular}}
	\vspace{-4mm}
\end{table}

\begin{table*}[t]
	\centering
	\caption{Ablation evaluation of the adaptive margin by self-mode-connectivity (SMC). The results with \textcolor{blue}{blue} color show how much the SMC improves, and the results with \textcolor{red}{red} color show the negative influence caused by SMC. \vspace{-2mm}
	}
	\label{tab_smc_performance}
	\resizebox{\textwidth}{!}{
		\setlength\tabcolsep{2.pt}
		\begin{tabular}{|c|c|cc|cc|cc|cc|}
			\toprule[1pt]
			\toprule[1pt]
			\multirow{3}{*} { \adjustbox{valign=b}{ \textbf{Datasets} } }  & \multirow{3}{*} {    \makecell[c]{\textbf{Unlearning} \\ \textbf{Sample Size}   } } & \multicolumn{2}{c}{\textbf{MIA($\%$)}} & \multicolumn{2}{c}{\textbf{RA ($\%$)}} & \multicolumn{2}{c}{\textbf{TA ($\%$)}} & \multicolumn{2}{c}{\textbf{RT} (second)}\\
			\cmidrule(lr){3-4}
			\cmidrule(lr){5-6}
			\cmidrule(lr){7-8}
			\cmidrule(lr){9-10}
			& & \makecell[c]{\textbf{ManiF} \\ Fixed Margin} & \makecell[c]{\textbf{ManiF-SMC} \\ Adaptive Margin}  
			& \makecell[c]{\textbf{ManiF} \\ Fixed Margin} & \makecell[c]{\textbf{ManiF-SMC} \\ Adaptive Margin}  
			& \makecell[c]{\textbf{ManiF} \\ Fixed Margin} & \makecell[c]{\textbf{ManiF-SMC} \\ Adaptive Margin}  
			& \makecell[c]{\textbf{ManiF} \\ Fixed Margin} & \makecell[c]{\textbf{ManiF-SMC} \\ Adaptive Margin}   \\
			\midrule[0.12em]
			\multirow{5}{*} {  \makecell[c]{\textbf{ \rotatebox{90}{ On MNIST }  } } }
			&200
			& 61.00  & 62.00  (\textcolor{blue}{$\uparrow$ 1.00})
			& 99.55  & 99.59 (\textcolor{blue}{$\uparrow$ 0.04})
			& 99.10  & 99.15 (\textcolor{blue}{$\uparrow$ 0.05})  
			& 1.33   & 1.48    (\textcolor{red}{$\uparrow$ 0.15})\\
			& 400    
			& 59.50 & 61.50 (\textcolor{blue}{$\uparrow$ 2.00})
			& 99.46  & 99.49 (\textcolor{blue}{$\uparrow$ 0.03})
			& 99.10  & 99.12 (\textcolor{blue}{$\uparrow$ 0.02})
			& 2.59 & 3.10 (\textcolor{red}{$\uparrow$ 0.51})   \\
			& 600   
			& 58.35  & 62.32 (\textcolor{blue}{$\uparrow$ 3.97})
			& 99.43  & 99.49 (\textcolor{blue}{$\uparrow$ 0.06})
			& 99.04  & 99.11 (\textcolor{blue}{$\uparrow$ 0.07}) 
			& 4.02  & 4.65 (\textcolor{red}{$\uparrow$ 0.63})  \\
			& 800 
			& 59.38  & 61.24 (\textcolor{blue}{$\uparrow$ 1.86})
			& 99.08  & 99.46 (\textcolor{blue}{$\uparrow$ 0.38})
			& 99.02  & 99.13 (\textcolor{blue}{$\uparrow$ 0.11}) 
			& 5.79  &6.78 (\textcolor{red}{$\uparrow$ 0.99}) \\
			& 1000    
			& 59.60  & 63.39 (\textcolor{blue}{$\uparrow$ 3.79})
			& 98.93  & 99.43 (\textcolor{blue}{$\uparrow$ 0.50})
			& 98.87  & 99.11 (\textcolor{blue}{$\uparrow$ 0.24})
			& 6.53  & 7.56 (\textcolor{red}{$\uparrow$ 1.03}) \\
			\midrule[0.12em]
			\multirow{5}{*} { \makecell[c]{  \textbf{ \rotatebox{90}{  On CIFAR10 }  } } }
			& 200
			& 56.00 & 59.00 (\textcolor{blue}{$\uparrow$ 3.00})
			& 98.96 & 99.04 (\textcolor{blue}{$\uparrow$ 0.08})
			& 81.71  & 81.75 (\textcolor{blue}{$\uparrow$ 0.04}) 
			& 2.02 & 2.37 (\textcolor{red}{$\uparrow$ 0.35}) \\
			& 400     
			& 58.78   & 63.07 (\textcolor{blue}{$\uparrow$ 4.29})
			& 98.50  &  98.46 (\textcolor{red}{$\downarrow$ 0.04})
			& 80.97  & 80.92 (\textcolor{red}{$\downarrow$ 0.05}) 
			& 4.52   & 5.19 (\textcolor{red}{$\uparrow$ 0.67}) \\
			& 600    
			&  63.19   & 66.98 (\textcolor{blue}{$\uparrow$ 3.79})
			& 97.16  & 97.68 (\textcolor{blue}{$\uparrow$ 0.52})
			& 79.60  & 80.25 (\textcolor{blue}{$\uparrow$ 0.65})
			&  6.40  &  7.09  (\textcolor{red}{$\uparrow$ 0.69})  \\
			& 800    
			& 63.89   & 67.88  (\textcolor{blue}{$\uparrow$ 3.99})
			& 96.12  & 97.14 (\textcolor{blue}{$\uparrow$ 1.02})
			& 78.10  & 79.09 (\textcolor{blue}{$\uparrow$ 0.99}) 
			& 8.57  & 9.32  (\textcolor{red}{$\uparrow$ 0.75}) \\
			& 1000   
			& 65.33   & 69.47 (\textcolor{blue}{$\uparrow$ 4.14})
			& 89.80  & 94.91 (\textcolor{blue}{$\uparrow$ 5.11})
			& 72.34  & 77.00 (\textcolor{blue}{$\uparrow$ 4.66})
			& 10.25  & 11.83 (\textcolor{red}{$\uparrow$ 1.58})   \\
			\bottomrule[1pt]
		\end{tabular}
	} 
	\begin{tabbing}
		\textbf{MIA}: Membership Inference Attack; \textbf{RA}: Remaining Accuracy; \textbf{TA}: Testing Accuracy; \textbf{RT}: Runing Time; \\ 
	\end{tabbing}
	\vspace{-4mm}
\end{table*}

\subsection{Application: Can ManiF-SMC Unlearn Generative Models?} \label{Generative_eval}

 \noindent
\textbf{Setup.} 
Generative tasks are also a key category of machine-learning services, so we evaluate ManiF-SMC in a generative learning setting. 
Specifically, we deploy ManiF-SMC to unlearn a VAE model \cite{kingma2014auto} and evaluate its performance with three metrics: the MIA accuracy to quantify unlearning effectiveness, and the mean-squared reconstruction error on the remaining (R-MSE) and test (T-MSE) sets to measure utility. The corresponding results for MNIST and CIFAR10 are shown in \Cref{additional_exp_on_black_b}. 


\noindent
\textbf{Results.}
In \Cref{additional_exp_on_black_b}, the MIA metric exhibits a clear upward trend when \textit{USS} increases on both MNIST and CIFAR10, demonstrating an effective unlearning of ManiF-SMC. Both R-MSE and T-MSE increase compared to the original model, showing the effective unlearning of ManiF-SMC while slight degradations of model utility, and from the observation of generated samples, the generative VAE model still keeps high-quality sample recovery after unlearning. 

\subsection{Application: Unlearning when Having Limited Access to Task Label Information} \label{access_limitation}

There are practical scenarios like semantic communications, which train encoders and decoders jointly by sender and receiver \citep{huang2022toward}. Unlearning requirements would be common for the sender but are impractical to implement as the sender can only access the encoder. Existing unlearning methods typically need access to encoder and decoder and training task information to design corresponding knowledge removal methods, commonly having the gradient ascent term \citep{guo2019certified}. Here, we conducted experiments to evaluate the applications of different methods in semantic communication scenarios when with or without the accessibility of the decoder. 

We present the results of unlearning semantic models trained on CelebA in \Cref{application_in_semantic_with_only_encoder}. If the decoder of the semantic communication models is not available, the existing unlearning methods (Retrain, GA, VBU, RFU) would be infeasible to implement unlearning for the semantic communication systems. They all need the training task information and access to the decoder as their unlearning methods and losses are highly related to the task. Only our ManiF-SMC is suitable for this complex but practical scenario, as ManiF-SMC can achieve unlearning based solely on the learned manifold representation, which only needs to access the encoder. Moreover, the ManiF-SMC also achieves the best performance similar to the retraining method with the decoder.

\begin{table}[t]
	\caption{Results of unlearning semantic communication systems with only access to encoder on CelebA. Only ManiF-SMC is suitable for this scenario when the semantic decoder is not available.
	}
	\vspace{-2mm}
	\label{application_in_semantic_with_only_encoder}
	\resizebox{\linewidth}{!}{
		\setlength\tabcolsep{7.5pt}
		\begin{tabular}{lccc}
			\toprule[0.8pt]
			\toprule[0.8pt]
			& MIA (\%)  & RA (\%) & TA (\%) \\
			\midrule
			Retrain (Decoder Available) & 60.00 & 96.59 & 96.13  \\  
			\rowcolor{verylightgray}
			Retrain (Dec. not Available) & -- & -- & --  \\  
			GA (Dec. Available) & 61.00 & 92.27 & 91.02 \\  
			\rowcolor{verylightgray}
			GA (Dec. not Avaliable) & -- & -- & --  \\ 
			VBU (Dec. Available) & 56.00 & 96.52 & 96.02  \\  
			\rowcolor{verylightgray}
			VBU (Dec. not Available) & -- & -- & --  \\ 
			RFU (Dec. Available) & 54.00 & 96.43 & 95.89  \\  
			\rowcolor{verylightgray}
			RFU (Dec. not Available) & -- & -- & --  \\ 
			SalUn (Dec. Available) & 54.50 & 96.51 & 96.27  \\  
			\rowcolor{verylightgray}
			SalUn (Dec. not Available) & -- & -- & --  \\ 
			ManiF-SMC (Dec. Available) & 54.50 & 96.34  & 95.93 \\  
			\rowcolor{verylightgray}
			ManiF-SMC (Dec. not Available) & 54.50 & 96.34  & 95.93  \\ 
			\bottomrule[0.8pt]
	\end{tabular}}
	\begin{tabbing}
		\hspace{2mm} --: the unlearning method is not applicable.
	\end{tabbing}
	\vspace{-4mm}
\end{table}

%% file: Contents/6_summary.tex

\section{Summary and Future Work} \label{s_a_fw}

In this paper, we investigate machine unlearning from a novel perspective: the model learned manifold representations. Motivated by our empirical observation, we first reformulate approximate unlearning as a representation-space relocation problem: \emph{push each erased sample away from its original representation while pulling it toward the centroid of its top-$k$ nearest retained samples.} We then propose ManiF-SMC, a label-agnostic unlearning method that operates purely in representation space. ManiF-SMC includes a manifold contrastive forgetting (ManiF) method to implement unlearning with a margin-based triplet loss and a self mode connectivity (SMC) method to adaptively calculate the margin.

We outline several promising directions stemming from this work. First, our experimental observation provides a new angle to reformulate a more robust approximate unlearning than previously gradient ascent-based perspectives. Second, our findings indicate that MMCRs can enhance existing unlearning approaches, offering useful design insights for future methods. Last but not least, investigating unlearning solely in the learned representation space provides broader applicability to more complex scenarios, such as pre-trained models or semantic communications, where a single encoder is shared across multiple downstream tasks. 

%% file: Contents/Appendix/appendix1.tex

\section{Training with MMCR for Generative Models} \label{t_mmcr}

\begin{figure} [t]
	\centering
	\hspace{-2mm}
	\includegraphics[width=0.52\linewidth]{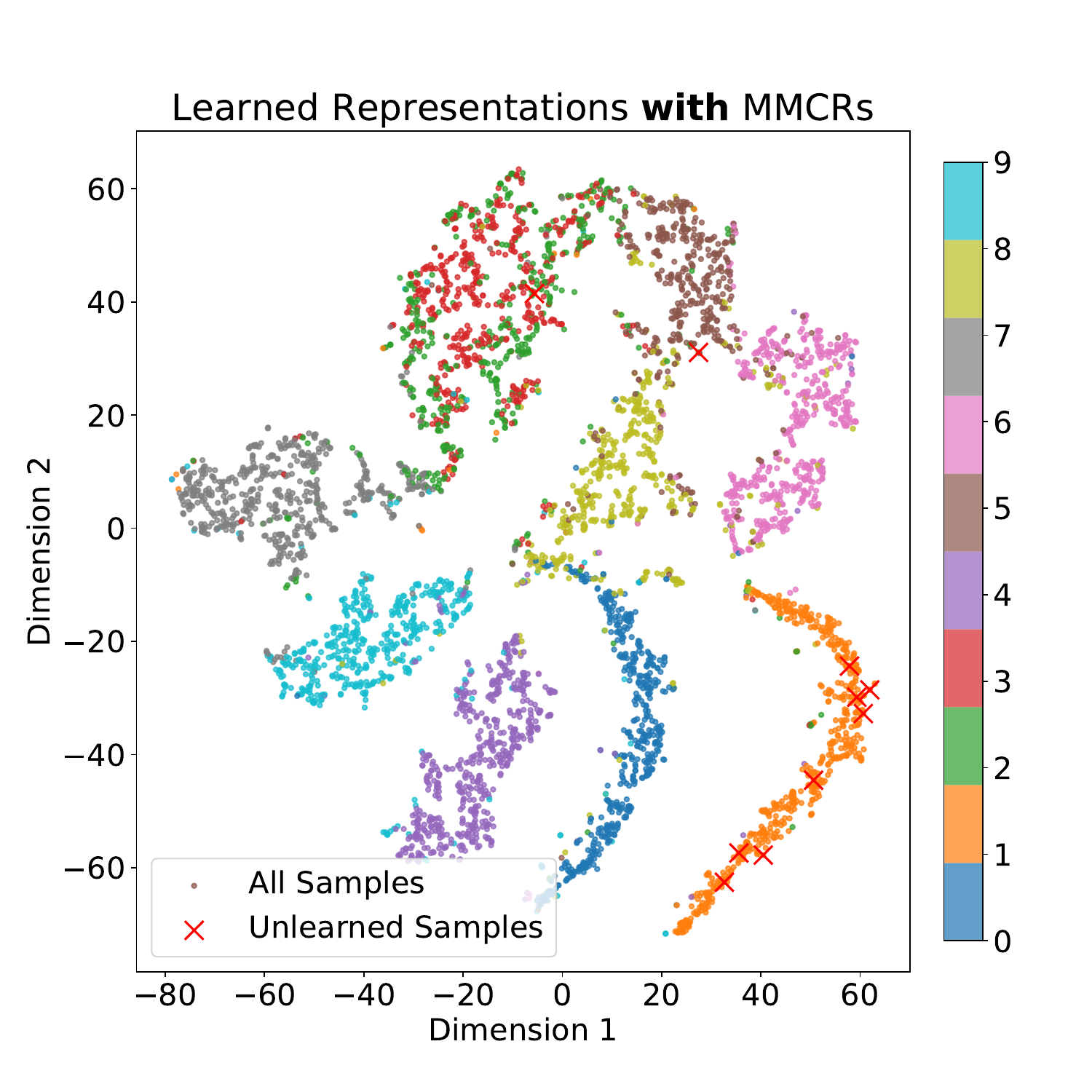}
	\hspace{-4mm}
	\includegraphics[width=0.52\linewidth]{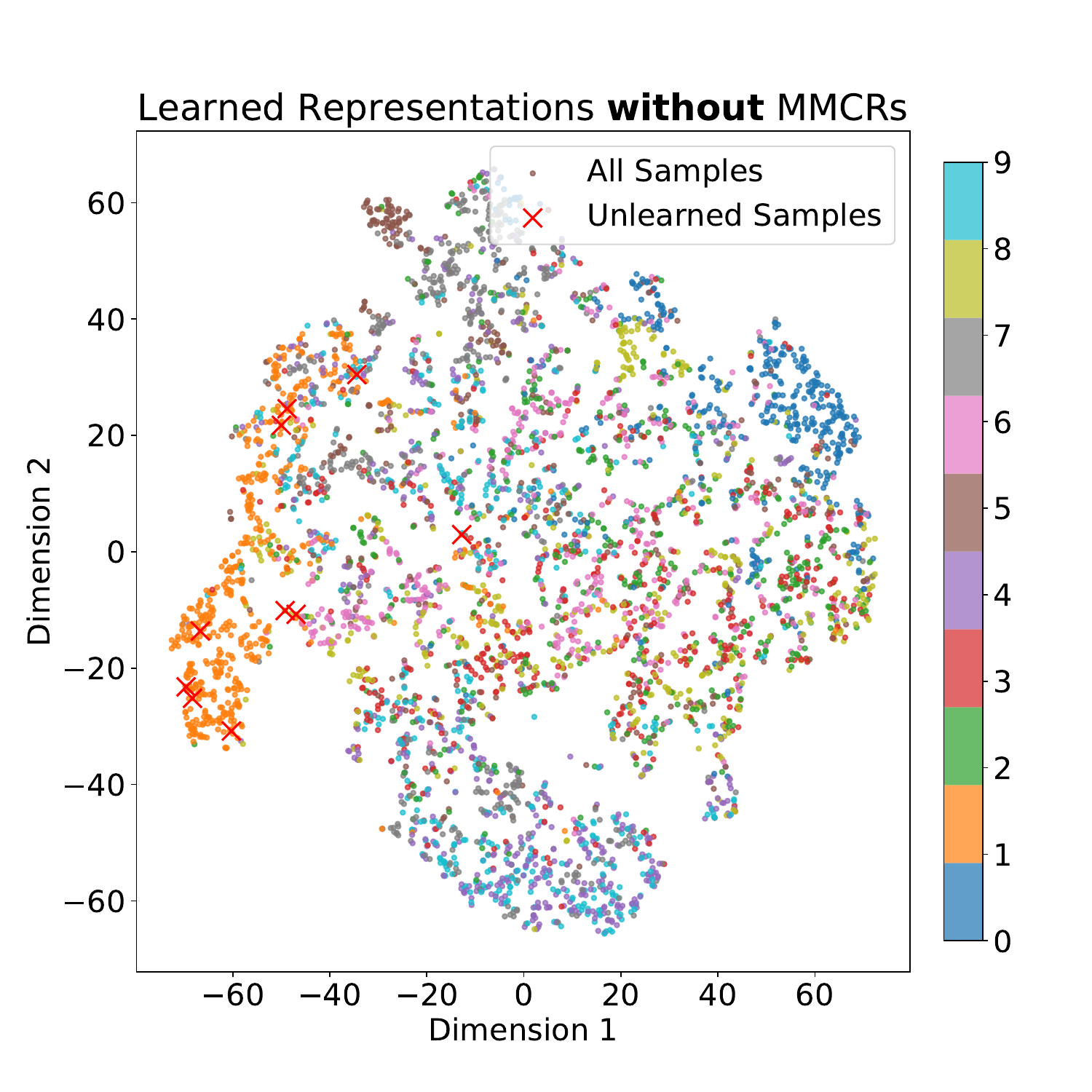}
	\vspace{-8mm}
	\caption{An example of models learned representation with or without MMCRs on MNIST. MMCRs support learning a separate representation, inspiring us to unlearn samples focusing only on the corresponding manifold representation. 
	} \vspace{-2mm}
	\label{representations_w_or_w_o_mmcrs}
\end{figure}

 \Cref{representations_w_or_w_o_mmcrs} demonstrates the comparison of training a self-supervise model with or without MMCRs \citep{yerxa2023learning} on MNIST. With MMCRs during model training, the representation will make every class's points stay close together (small intra-class distance) while different classes lie far apart (large inter-class distance). The representation with these properties inspires us to investigate an efficient approximate unlearning method within only the manifold representation. 
 
 
 \begin{table}[h]
 	\centering
 	\caption{Frequently used notations in ManiF-SMC. \vspace{-2mm}}
 	\label{tab:notation_manif_smc}
 	\footnotesize
 	\setlength{\tabcolsep}{6pt}
 	\renewcommand{\arraystretch}{1.0}
 	\begin{tabular}{@{}l p{0.385\textwidth}@{}}
 		\toprule
 		Symbol & Meaning \\
 		\midrule
 
 		$\theta_o$ & Original model parameters (trained on $S$).\\
 		$\theta_u$ & Unlearned model parameters after forgetting $S_u$.\\
 		$\tilde{\theta}$ & Surrogate model sampled on the SMC path, $\tilde{\theta}=\phi_w(t^\star)$.\\
 
 		$S$ & Original training set.\\
 		$S_u$ & Unlearning (erased) set.\\
 		$S_r$ & Remaining (retained) set, $S_r = S\setminus S_u$.\\
 		$x_i$ & An erased target sample, $x_i\in S_u$.\\
 		$S_k^i$ & Top-$k$ most similar retained samples for $x_i$, selected in the original representation space; $S_k^i\subseteq S_r$.\\
 		$S_k$ & Union of retained neighborhoods, $S_k=\cup_{x_i\in S_u} S_k^i$.\\
 		$k$ & Number of retained neighbors used for each erased sample.\\
 		$\texttt{z}_{i,o}$ & Original representation of $x_i$: $\texttt{z}_{i,o}=f_{\theta_o}(x_i)$.\\
 		$\texttt{z}_{i,u}$ & Unlearned representation of $x_i$: $\texttt{z}_{i,u}=f_{\theta_u}(x_i)$.\\
 		$\texttt{c}_{i,o}$ & Neighbor centroid under the original model (computed on $S_k^i$ using $f_{\theta_o}$).\\
 		$\texttt{c}_{i,u}$ & Neighbor centroid under the unlearned model: $\texttt{c}_{i,u}=\frac{1}{|S_k^i|}\sum_{x_j\in S_k^i} f_{\theta_u}(x_j)$.\\
 		$\texttt{c}_{i,\tilde{\theta}}$ & Neighbor centroid estimated by the surrogate model: $\texttt{c}_{i,\tilde{\theta}}=\frac{1}{|S_k^i|}\sum_{x_j\in S_k^i} f_{\tilde{\theta}}(x_j)$.\\

 		$w$ & Learnable control point of the quadratic B\'ezier path.\\
 		$t$ & Path parameter, sampled from $\mathcal{U}[0,1]$ when optimizing $w$.\\
 		$t^\star$ & Fixed sampling position on the path (e.g., $t^\star=0.5$) used to form $\tilde{\theta}$.\\

 		\bottomrule
 	\end{tabular}
 	\vspace{-2mm}
 \end{table}
 
 \section{Notations in ManiF-SMC} \label{notations_in_ManiF_SMC}
 
 We summarize the notations in \Cref{tab:notation_manif_smc}.

 \begin{algorithm}[h]
 	\caption{ManiF-SMC} \label{algManiF_SMC}
 	\begin{normalsize}
 		\BlankLine
 		\KwIn{Trained model $\theta_o$; unlearning set $S_u$; retained neighbor sets $\{S_k^i\}_{x_i\in S_u}$ and $S_k=\cup_{x_i\in S_u} S_k^i$; distance $\mathrm{dist}(\cdot,\cdot)$; control point $w$; learning rate $\eta$; epochs $E$; fixed $t^\star$ (e.g., $0.5$).}
 		\KwOut{Updated model parameters $\theta_u$.}
 		
 		\BlankLine
 		\tcp{Initialization}
 		$\theta_u \gets \theta_o$ \tcp*{start from original model}
 		$w \gets \theta$ \tcp*{initialize Bézier control point}
 		\ForEach{$x_i \in S_u$}{
 			$\texttt{z}_{i,o} \gets f_{\theta_o}(x_i)$ \tcp*{cache original representation}
 		}
 		
 		\BlankLine
 		\For{$e=1$ \KwTo $E$ epochs}{
 			\tcp{(A) Learn SMC control point $w$ on retained neighborhood data}
 			\ForEach{mini-batch $B_k \subset S_k$}{
 				sample $t \sim \mathcal{U}[0,1]$\;
 				$\theta_t \gets \phi_w(t) = (1-t)^2\theta_u + 2t(1-t)w + t^2\theta_o$\;
 				$\mathcal{L}_{\mathrm{path}} \gets \mathcal{L}_{B_k}(\theta_t)$ \tcp*{retained loss (same as training loss)}
 				$w \gets w - \eta \nabla_w \mathcal{L}_{\mathrm{path}}$\;
 			}
 			
 			\BlankLine
 			\tcp{(B) Pick a surrogate model on the low-loss path}
 			$\tilde{\theta} \gets \phi_w(t^\star)$ \tcp*{we pick $t^\star=0.5$ in experiments} 
 			
 			\BlankLine
 			\tcp{(C) ManiF update: compute centroid + adaptive margin under $\tilde{\theta}$, update $\theta_u$}
 			\ForEach{mini-batch $B_u \subset S_u$}{
 				\ForEach{$x_i \in B_u$}{
 					$\texttt{c}_{i,\tilde{\theta}} \gets \frac{1}{|S_k^i|}\sum_{x_j \in S_k^i} f_{\tilde{\theta}}(x_j)$\;
 					$\alpha_i \gets \Big[\mathrm{dist}\big(f_{\tilde{\theta}}(x_i),\texttt{z}_{i,o}\big) - \mathrm{dist}\big(f_{\tilde{\theta}}(x_i),\texttt{c}_{i,\tilde{\theta}}\big)\Big]_+$\;
 					$\ell_i \gets \Big[\mathrm{dist}\big(f_{\theta_u}(x_i),\texttt{c}_{i,\tilde{\theta}}\big) - \mathrm{dist}\big(f_{\theta_u}(x_i),\texttt{z}_{i,o}\big) + \alpha_i\Big]_+$\;
 				}
 				$\mathcal{L}_{\mathrm{triplet}} \gets \sum_{x_i \in B_u} \ell_i$\;
 				$\theta_u \gets \theta_u - \eta \nabla_{\theta_u}\mathcal{L}_{\mathrm{triplet}}$\;
 			}
 		}
 		\BlankLine
 		\Return $\theta_u$\;
 	\end{normalsize}
 \end{algorithm}

\section{Implementation Details of ManiF-SMC and Discussion} \label{algorithm_appendix}

\textbf{Overview.}
Algorithm~\ref{algManiF_SMC} summarizes ManiF-SMC, which alternates between (i) constructing a retained-geometry surrogate via self mode connectivity (SMC) and (ii) updating the unlearning model parameters via the manifold contrastive forgetting (ManiF) objective. The key design is that both the retained-neighbor centroid and the triplet margin are computed under a surrogate model sampled on a low-loss path trained only on retained neighborhood data, avoiding the need for retraining-from-scratch on $S_r$.

\paragraph{Inputs and preprocessing.}
The algorithm takes the trained model $\theta_o$, the erased (unlearning) set $S_u$, and for each $x_i\in S_u$ a retained neighborhood $S_k^i$ (top-$k$ most similar samples from the remaining data), with $S_k=\cup_{x_i\in S_u} S_k^i$. In practice, $S_k^i$ can be built once using the original representation space (e.g., nearest neighbors of $\texttt{z}_{i,o}=f_{\theta_o}(x_i)$) and then kept fixed during unlearning to reduce overhead.
We also cache $\texttt{z}_{i,o}$ for all erased samples (Lines 3--4), since it is used as the ``negative'' reference in the triplet constraint and does not change throughout the process.

We should notice that this algorithm does not rely on the label information. Hence, our inputs do not include the $y_i$.

\paragraph{Step (A): learning the SMC path control point $w$.}
Given endpoints $\theta_u$ (current unlearning model) and $\theta_o$ (original model), we parameterize a quadratic B\'ezier curve
$\phi_w(t) = (1-t)^2\theta_u + 2t(1-t)w + t^2\theta_o$.
In Step (A), we update the control point $w$ by minimizing the retained loss along the path (Eq.~\eqref{eq:w_opt}) using only mini-batches from $S_k$ (Lines 5--10). This yields a low-loss connection in parameter space that approximates the retained-data geometry without requiring full retraining on $S_r$. Importantly, the gradient is taken \emph{w.r.t. $w$ only}; $\theta_u$ is held fixed during this inner update so that the path training remains lightweight.

\paragraph{Step (B): sampling a surrogate model $\tilde{\theta}$.}
After optimizing $w$ on the retained neighborhood, we select a surrogate model $\tilde{\theta}=\phi_w(t^\star)$ (Line 11). We use a fixed $t^\star$ (e.g., $0.5$) for simplicity and stability. Empirically, mid-path sampling often provides a good trade-off between staying close to the retained geometry and not overfitting to either endpoint, while preserving the ``low-loss'' property by construction.

\paragraph{Step (C): ManiF update with centroid and adaptive margin.}
For each erased sample $x_i$, we compute the retained-neighbor centroid under the surrogate model,
$\texttt{c}_{i,\tilde{\theta}}=\frac{1}{|S_k^i|}\sum_{x_j\in S_k^i} f_{\tilde{\theta}}(x_j)$,
and an adaptive margin
$\alpha_i=\big[\mathrm{dist}(f_{\tilde{\theta}}(x_i),\texttt{z}_{i,o})-\mathrm{dist}(f_{\tilde{\theta}}(x_i),\texttt{c}_{i,\tilde{\theta}})\big]_+$
(Lines 14--15), which corresponds to Eq.~\eqref{adaptive_margin}. We then update $\theta_u$ by minimizing the triplet hinge objective (Eq.~\eqref{eq_triplet_loss_with_function}) on mini-batches from $S_u$ (Lines 17--18). Conceptually, this enforces that the updated representation $f_{\theta_u}(x_i)$ moves at least $\alpha_i$ closer to the retained-neighbor centroid than to its original representation $\texttt{z}_{i,o}$, implementing the push--pull unlearning behavior in a single margin-ranking constraint.

\paragraph{Why the surrogate is necessary.}
A core difficulty in representation-based unlearning is that the desired centroid $\texttt{c}_{i,u}$ and a suitable margin $\alpha$ are naturally defined under a retrained model on $S_r$, which is unavailable in efficient approximate unlearning. ManiF-SMC addresses this by using SMC to construct $\tilde{\theta}$ that is trained only on retained neighborhood data. Since $\tilde{\theta}$ is sampled from a low-loss connecting path, it provides a stable estimate of retained-neighbor representations (supported by Proposition~\ref{prop:smc_logit2}) and yields a data-dependent margin $\alpha_i$ that adapts to local geometry rather than using a global constant.

\paragraph{Distance function choices.}
We use Euclidean distance by default for $\mathrm{dist}(\cdot,\cdot)$ in Algorithm~\ref{algManiF_SMC}, which is standard in metric learning. Alternative metrics (e.g., cosine distance and $L_2$ norm) can be plugged in without changing the algorithm structure. We provide some evaluations about distance metrics in \Cref{distance_metrics_eval}.


\paragraph{Other hyperparameters.}
Neighborhood size $k$ and sampling $t$ of SMC will also influence the performance of \Cref{algManiF_SMC}. We also provide the additional evaluations in \Cref{add_top_k_ablation,t_eval}.

\paragraph{Complexity discussion.}
Compared with retraining on $S_r$, ManiF-SMC is efficient because it (i) computes centroids only over the retained neighborhoods $S_k^i$ rather than the full retained set, and (ii) optimizes the SMC path using only $S_k$. The dominant additional cost beyond the ManiF update is the forward/backward pass for path training of $w$ on $S_k$ and the centroid computation for each $x_i$ within a mini-batch. Both are controllable via $k$ and the number of SMC optimization steps.

 \begin{table}[h]
	\scriptsize
	\caption{Dataset statistics.} 
	\label{dataset_table}
	\resizebox{\linewidth}{!}{
		\setlength\tabcolsep{7.5pt}
		\begin{tabular}{cccc}
			\toprule[0.8pt]
			Dataset & Feature Dimension  & \#. Classes & \#. Samples \\
			\midrule
			MNIST \citep{deng2012mnist} & 28×28×1 & 10 & 70,000  \\  
			\rowcolor{verylightgray}
			CIFAR10 \citep{krizhevsky2009learning} & 32×32×3 & 10 & 60,000  \\  
			CelebA~\citep{liu2018large} & 178×218×3 & 2 (Gender) & 202,599 \\
			\rowcolor{verylightgray}
			Tiny-ImageNet \citep{le2015tiny} & 64×64×3 & 200 & 110,000 \\
			\bottomrule[0.8pt]
	\end{tabular}}
\vspace{-2mm}
\end{table}

\section{Datasets} \label{datasets_appendix}

The statistics of all datasets used in our experiments are listed in \Cref{dataset_table}. Both MNIST and CIFAR10 are used to train 10-class classification models. The experiment on CelebA is to identify the gender attributes of the face images. The task is a binary classification problem, different from the ones on MNIST and CIFAR10. The task of Tiny-ImageNet is a 200-class classification. These datasets offer a range of objective categories with varying levels of learning complexity. We also introduce them as below.


\begin{itemize}[itemsep=0pt, parsep=0pt, leftmargin=*]
	\item \textbf{MNIST~\citep{deng2012mnist}.} MNIST contains 60,000 handwritten digit images for the training and 10,000 handwritten digit images for the testing. All these black and white digits are size normalized, and centered in a fixed-size image with 28 × 28 pixels.
	\item \textbf{CIFAR10~\citep{krizhevsky2009learning}.} CIFAR10 dataset consists of 60,000 32x32 colour images in 10 classes, with 6,000 images per class. There are 50,000 training images and 10,000 test images.
	\item  \textbf{CelebA~\citep{liu2018large}.} CelebA is a large-scale face attributes dataset with more than 200,000 celebrity images, each with 40 attribute annotations.
	\item \textbf{Tiny-ImageNet~\citep{le2015tiny} .} The Tiny-ImageNet image size is 64x64 pixels and the dataset sizes are 100,000 training images across 200 classes; 10,000 test images.
\end{itemize}

 \begin{figure*}[t]
	\centering
	\hspace{-2mm}
	\subfloat{  \label{fig:celebamodelmiauss}  \rotatebox{90}{ \hspace{12mm}	\small{On CelebA} }
		\includegraphics[scale=0.3]{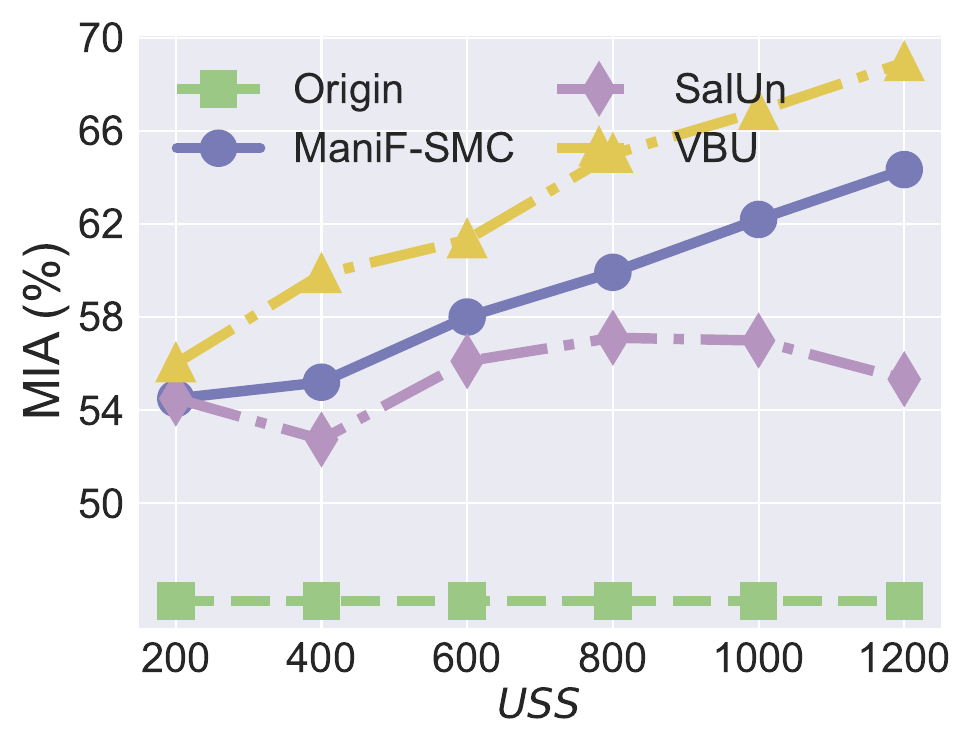}
	}	
	\hspace{-2mm}
	\subfloat{ 	 	\label{fig:celebamodelrauss}
		\includegraphics[scale=0.3]{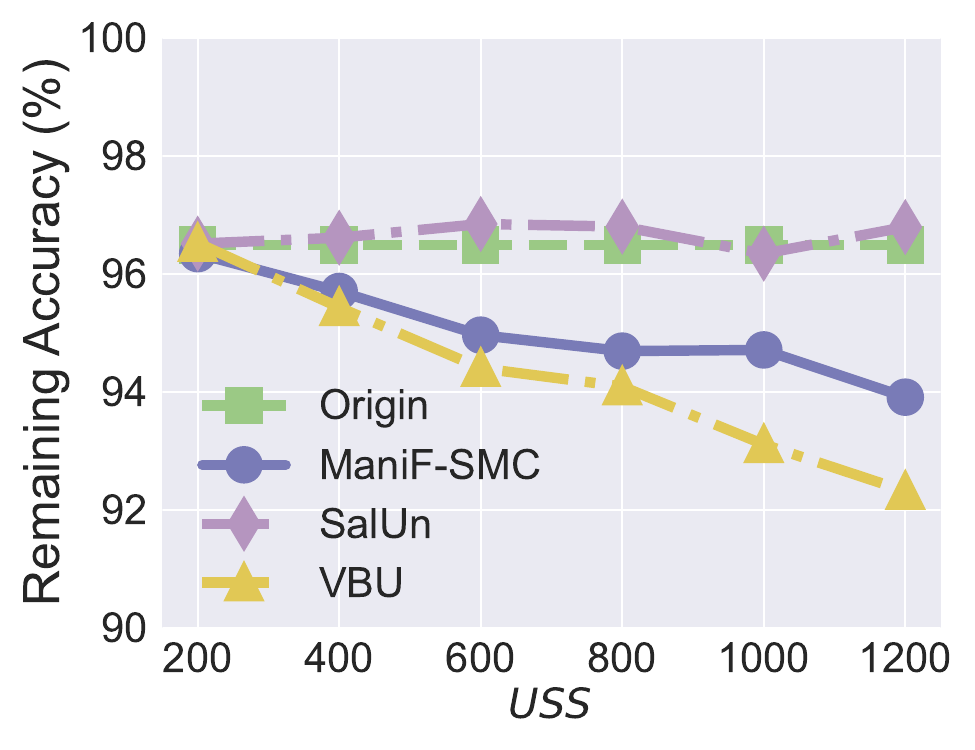}
	}
	\hspace{-2mm}
	\subfloat{  		\label{fig:celebamodeltauss}
		\includegraphics[scale=0.3]{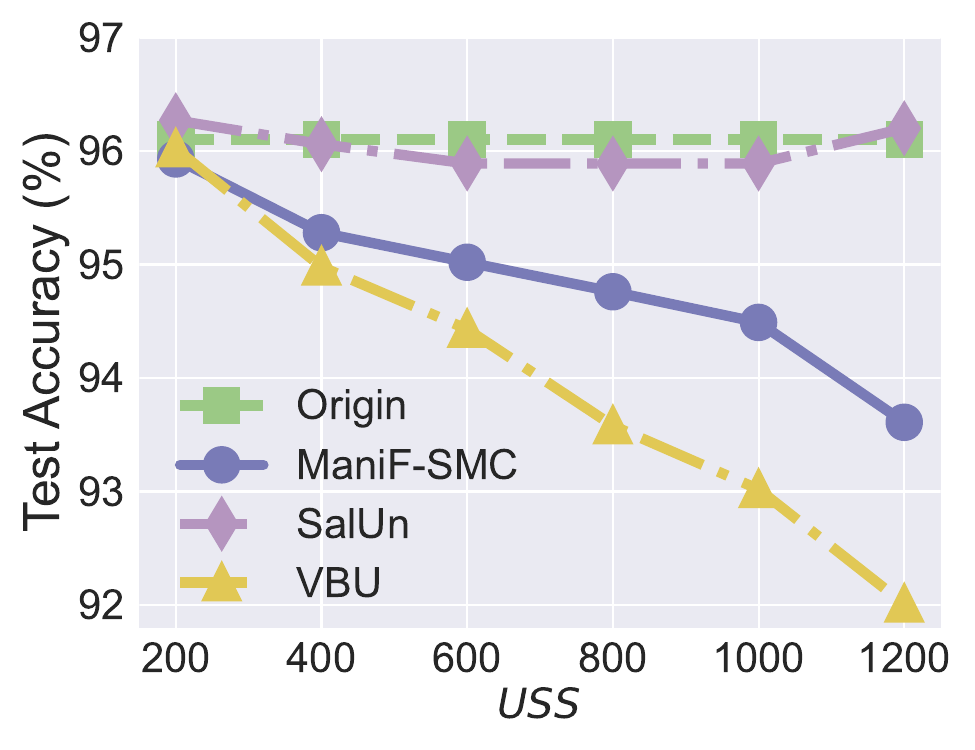}
	}
	\caption{Additional Evaluations of impact about different $\it USS$ on various unlearning methods on CelebA.   } 
	\label{evaluation_of_uss_cifar10_celeba} 
\end{figure*}

\section{Evaluation Metrics} \label{metric_detail}
We provide the details of metrics as follows.

\begin{itemize}[itemsep=0pt, parsep=0pt, leftmargin=*] 
	\item \textbf{Membership inference attack (MIA).} We employ the membership inference attack (MIA) \citep{song2019privacy,yeom2018privacy} to gauge the effectiveness of unlearning. Concretely, we apply the MIA predictor to the unlearned model $f_{\theta_u}$ on the dataset $S_u$. The resulting success rate of MIA is calculated by how many instances in $S_u$ are correctly identified as non-training samples for $f_{\theta_u}$. A higher MIA success rate suggests that $f_{\theta_u}$ retains less information about $S_u$.
	\item  \textbf{Remaining accuracy (RA).} This refers to the accuracy of $f_{\theta_u}$ on the remaining dataset $S_r$, which reflects the fidelity of machine unlearning. The training data information should be preserved from original model to the unlearned model.
	\item \textbf{Testing accuracy (TA).} We report TA as an indicator of how well the unlearned model $f_{\theta_u}$ generalizes when evaluated on the test dataset.
		\item  \textbf{Remaining mean-squared error (R-MSE).} This refers to the reconstruction mean-squared error of generative models on the remaining dataset $S_r$, which reflects the fidelity of machine unlearning for generative models.
	\item \textbf{Testing mean-squared error (T-MSE).} We report T-MSE as an indicator of how well the unlearned generative model performs when evaluated on the test dataset.
	\item  \textbf{Running Time (RT).} This metric reflects the computational efficiency of a machine unlearning method. We obtain RT by tracking the per-batch training duration and multiplying by the total number of training epochs.
\end{itemize}

\section{Machine Unlearning Benchmarks} \label{revisiting_mu}

Machine unlearning seeks to remove or negate the influence of a subset of training data $\{x_i\}_{i \in S_u}$ on a trained model $f_{\theta}$ \citep{bourtoule2021machine,warnecke2024machine}. Although exact unlearning can be achieved through retraining a model using the remaining dataset, the associated computational costs have driven the more efficient solutions, the approximate unlearning methods \citep{shen2024labelagnostic,izzo2021approximate}. Classic approximate unlearning approaches include:

\begin{itemize}[itemsep=0pt, parsep=0pt, leftmargin=*]
	\item \textit{Gradient Ascent (\textbf{GA})} \citep{graves2021amnesiac,thudi2022unrolling}: GA reverses the model training on the erased samples $S_u$ by adding the corresponding gradients back to $\theta_o$, i.e., moving $\theta_o$ in the direction of increasing loss for data to be unlearned, where $\theta_o$ is the original trained model parameters.
	\item \textit{Variational Bayesian Unlearning (\textbf{VBU})} \citep{nguyen2020variational,nguyen2022markov}: VBU is an approximate unlearning method based on variational Bayesian inference. In practice, a middle layer of original neural networks is used as the Bayesian layer to calculate the VBU loss according to \citep{nguyen2020variational} to achieve unlearning.
	\item  \textit{Representation Forgetting Unlearning \textbf{(RFU)}} \citep{wang2023machine}: RFU tries to unlearn a bottleneck representation by minimizing the mutual information between the representation and the erased samples, which provides an unlearning solution from an information theory perspective. 
	\item  \textit{Saliency Unlearning \textbf{(SalUn)}} \citep{fan2024salun}: SalUn introduces ``weight saliency'' to remove the influence of samples and classes for unlearning a model, improving effectiveness and efficiency.
\end{itemize}

From the manifold representation viewpoint, unlearning a sample $(x_i,y_i) \in S_u$ can be achieved by pushing its representation $\texttt{z}_i$ away from the center of its learned local manifold representation, for example, the class-manifold centroid $\texttt{c}_{y_i}$. One simple operation proxy is to maximize the distance $\| \texttt{z}_i - \texttt{c}_{y_i} \|$, ensuring that the unlearning sample no longer ``lives'' near its old class manifold centroid. 



\section{Additional Evaluations}

\subsection{Additional Efficiency Evaluations about \textit{USS}} \label{add_exp_eff_uss}

We put the additional unlearning effectiveness evaluation on CelebA in \Cref{evaluation_of_uss_cifar10_celeba}, which shows the data removal effect and model utility preservation of ManiF-SMC.

\begin{figure}[t]
	\centering
	\subfloat[\small On CelebA]{ 		\label{fig_celebarunningtimecsasbar}
		\includegraphics[scale=0.32]{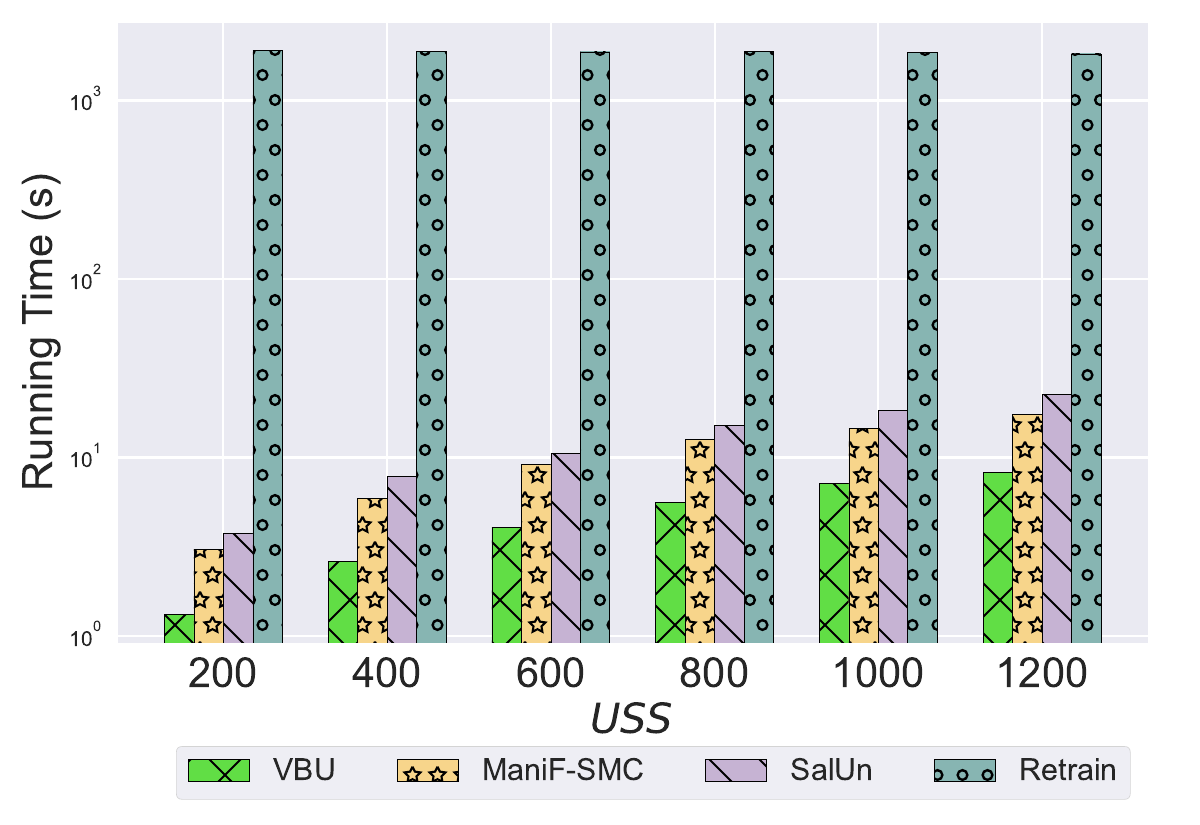}
	}
	\vspace{-2mm}
	\caption{Efficiency of Running time. Although the RT of approximate unlearning methods (VBU, ManiF-SMC, and SalUn) increases as the \textit{USS} increases, it is still much more efficient than retraining. \vspace{-0mm} } 
	\label{additional_efficiency_of_uss} 
	\vspace{-4mm}
\end{figure}

We put the additional efficiency evaluation results on CelebA, in \Cref{additional_efficiency_of_uss}. The running-time curves show a clear efficiency hierarchy: the exact Retrain baseline is always the slowest—often by two to three orders of magnitude—while the approximate unlearning methods (VBU, ManiF, and SalUn) finish far sooner. Among the approximations, VBU is consistently the fastest, SalUn incurs the highest overhead, and ManiF-SMC sits in between, mirroring the computational complexity built into their respective update rules. 

\begin{figure*}[t]
	\centering
	\subfloat{		\label{fig:mnistmodelmiaussfinetuned} \rotatebox{90}{ \hspace{14mm}	\scriptsize{On MNIST} }
		\includegraphics[scale=0.3]{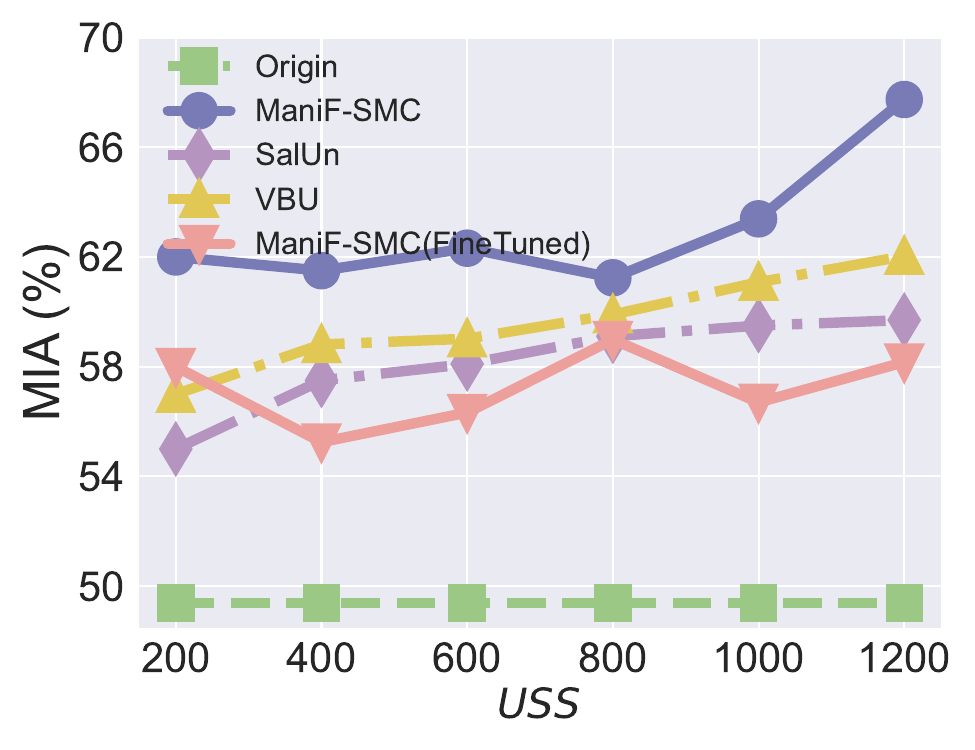}
	}			
	\subfloat{ 		\label{fig:mnistmodelraussfinetuned}
		\includegraphics[scale=0.3]{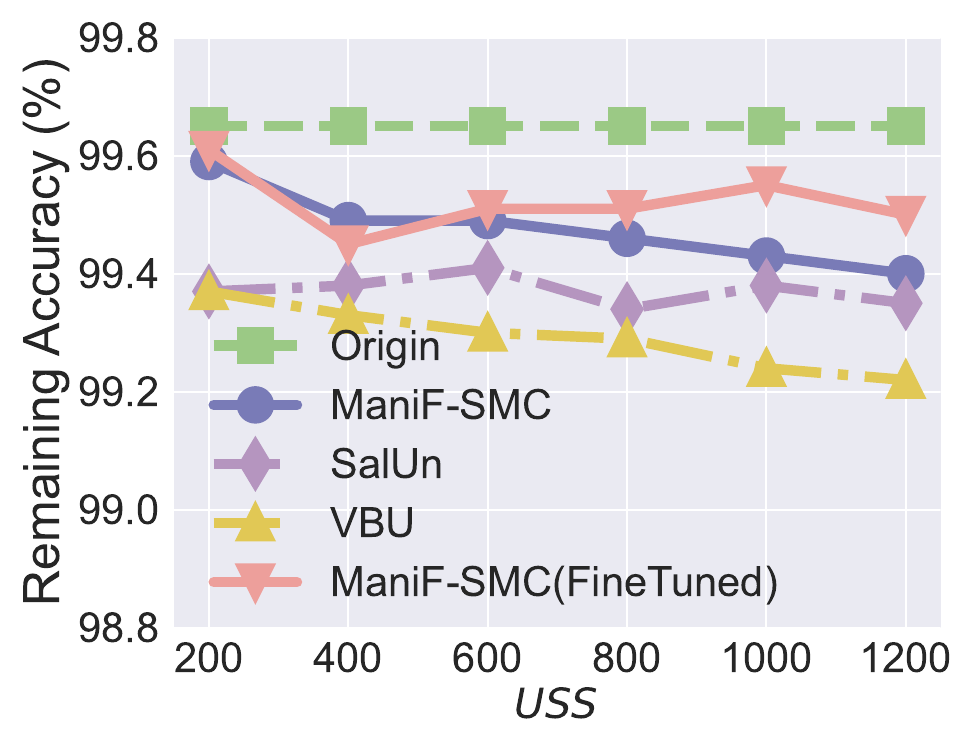}
	}
	\subfloat{  		\label{fig:mnistmodeltaussfinetuned}
		\includegraphics[scale=0.3]{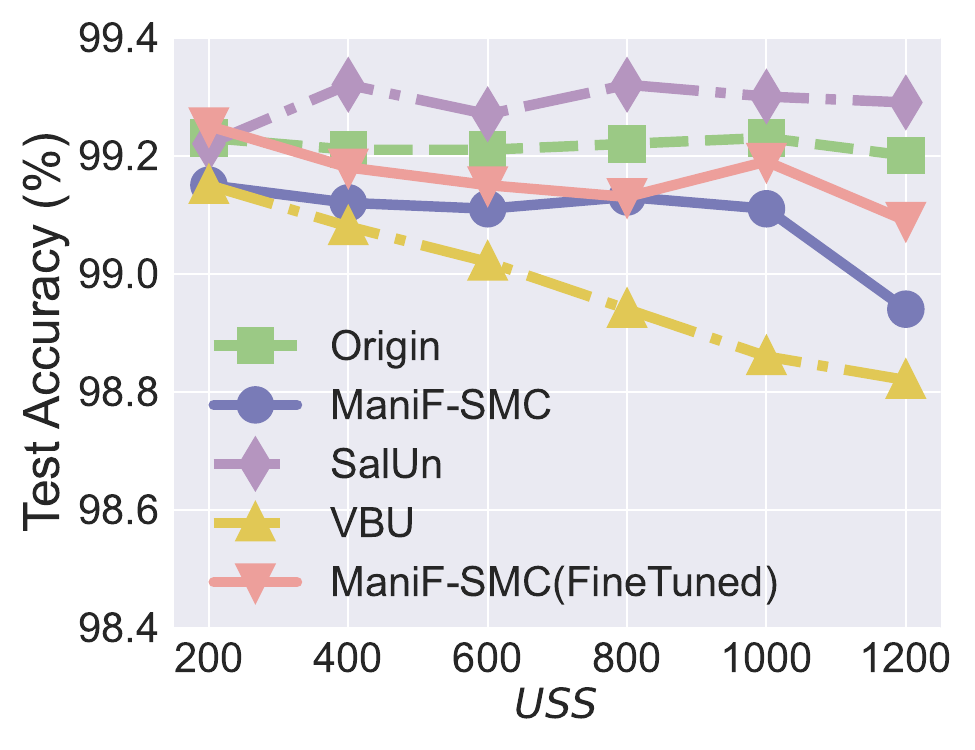}
	}
	\vspace{-2mm} \\
	\subfloat{	\label{fig:cifar10modelmiaussfinetuned} \rotatebox{90}{ \hspace{14mm}	\scriptsize{On CIFAR10} }
		\includegraphics[scale=0.3]{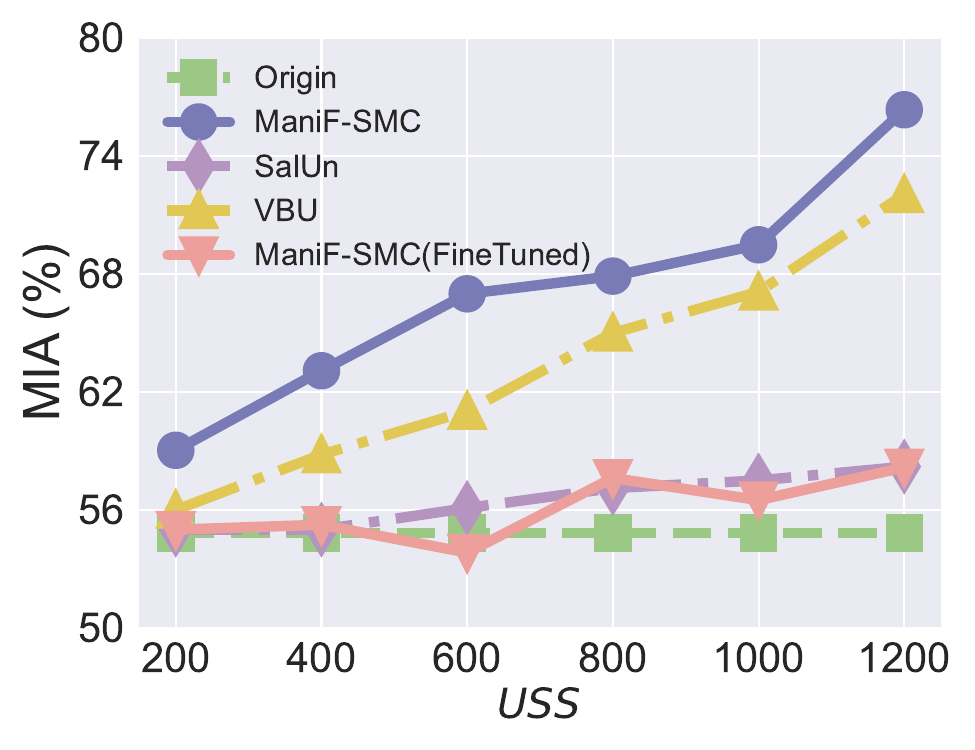}
	}			
	\subfloat{ 		\label{fig:cifar10modelraussfinetuned}
		\includegraphics[scale=0.3]{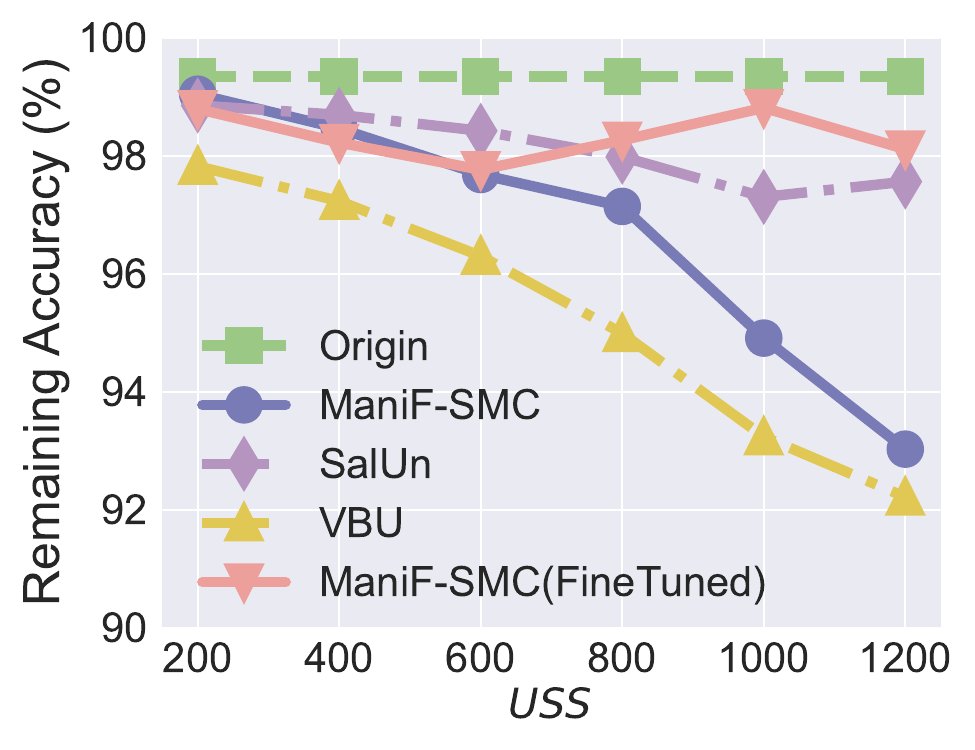}
	}
	\subfloat{  		\label{fig:cifar10modeltaussfinetuned}
		\includegraphics[scale=0.3]{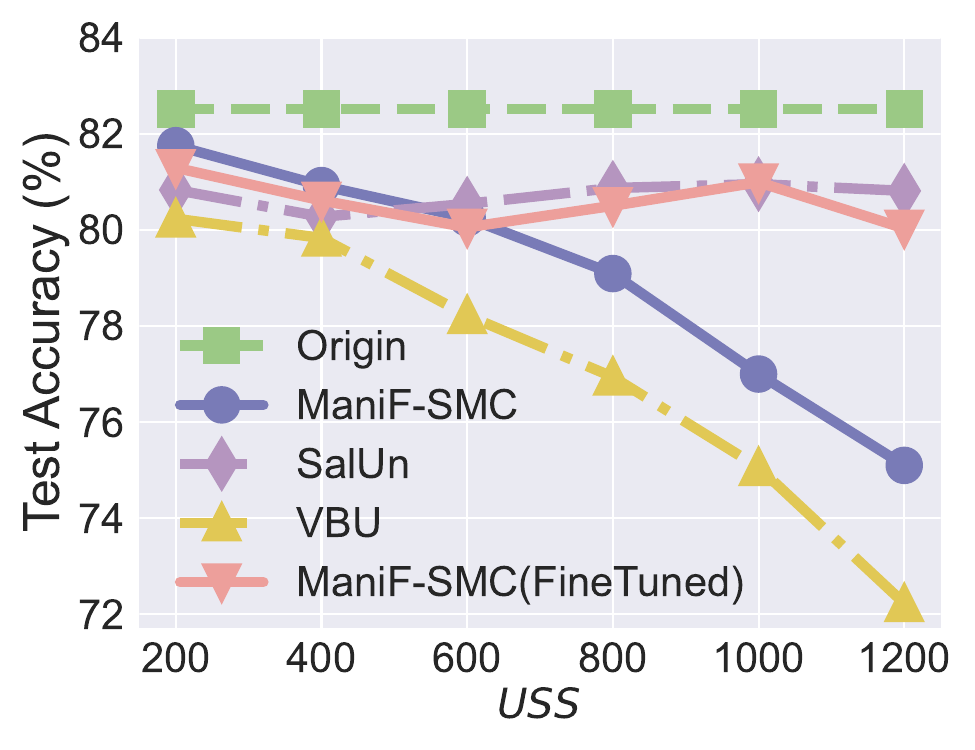}
	}
	\caption{Additional evaluations for ManiF-SMC with fine-tuning utilizing the remaining dataset. ManiF-SMC with fine-tuning using the remaining dataset can effectively mitigate the model utility (RA and TA) degradation during unlearning, meanwhile the unlearning effectiveness (MIA) has slight drops compared with ManiF-SMC without fine-tuning. \vspace{-0mm} } 
	\label{evaluation_of_mcr_r} 
\end{figure*}

\subsection{Limitations and Additional Evaluation with Fine Tuning} \label{with_finetuning}

\noindent
\textbf{Limitations of ManiF-SMC.} Since ManiF-SMC unlearns models only using the model manifold representation without the data task information, ManiF-SMC suffers greater utility loss compared with SalUn and other approximate unlearning methods that utilize the remaining data to fine-tune the unlearned model, especially when the unlearning sample size is large. Mitigating this degradation is therefore an important avenue for future research. In this work, we introduce a simple fine-tuning step on the remaining dataset and assess its effect on model performance.

\noindent
\textbf{Setup.} We additionally test the improvement of fine tuning for ManiF-SMC on MNIST and CIFAR10. The \textit{USS} is set from 200 to 1200, where 1200 is around $2\%$ training data on MNIST and CIFAR10. Our other settings are the same as the previous test for ManiF-SMC. The corresponding results are presented in \Cref{evaluation_of_mcr_r}. 
 
\noindent
\textbf{Evaluation Results.} In \Cref{evaluation_of_mcr_r},  we observed that ManiF-SMC with fine-tuning using the remaining dataset can effectively mitigate the model utility (RA and TA) degradation during unlearning, meanwhile the unlearning effectiveness (MIA) has slight drops compared with ManiF-SMC without fine-tuning. We should notice that the original ManiF-SMC unlearns only based on the model learned manifold representation, not needing the task information. Adding the fine-tuning on the remaining dataset will need label information to calculate the original learning loss. It improves the model utility preservation but makes the ManiF-SMC similar to the existing unlearning methods, having not cut the reliance on the task information.

\subsection{Ablation Study: How do Distance Metrics Influence the Performance of ManiF-SMC?} \label{distance_metrics_eval}

\noindent
\textbf{Setup.} 
As we mentioned above, different distance metrics may influence the unlearning methods. 
Therefore, we conducted the experiments using other metrics, the cosine similarity (the Normalized Temperature-Scaled Cross Entropy (NT-Xent) used in SimCLR \cite{chen2020simple}). For a better adaption of cosine similarity (NT-Xent), we also conducted Cosine Similarity (NT-Xent) with Multi-positive samples (add an additional one), because our method is for unlearning, the negative samples are limited.

\noindent
\textbf{Results.}
In \Cref{Different_Distance_Metrics}, the distance using cosine similarity achieves similar model utility preservation as the $L_2$-norm. However, the cosine similarity achieves a lower MIA value for unlearning effect than $L_2$-norm. We infer when using similarity as the triplet distance, the similarity has a higher preference to optimize the (anchor, positive) pair, making unlearning more difficult than the $L_2$-norm. For Cosine Similarity (NT-Xent) with Multi-positive samples, it does increase the unlearning model utility but further reduces the unlearning effect, the MIA. Moreover, we conducted experiments with the label information for fine tuning: $L_2$-norm (positive with label). It increases the unlearned model utility too and shows the potential of fine tuning for ManiF-SMC. However, the biggest advantage of our method is still not relying on the task and corresponding gradient ascent.

\begin{table}[t]
	\caption{Evaluation of Different Distance Metrics for ManiF-SMC on MNIST. 
	}
	\vspace{-2mm}
	\label{Different_Distance_Metrics}
	\resizebox{\linewidth}{!}{
		\setlength\tabcolsep{5.5pt}
		\begin{tabular}{ccccc}
			\toprule[0.8pt]
			\toprule[0.8pt]
			Distance Metric & MIA (\%)  & RA (\%) & TA (\%) & RT (second) \\
			\midrule
			$L_2$ norm &62.00 	& 99.59 	& 99.15 	&1.482 \\  
			\rowcolor{verylightgray}
			$L_2$ norm (with label) & 58.00  & 99.62 & 99.20 & 1.598  \\   
			Cosine Similarity (NT-Xent) &60.00 	& 99.38 &	99.04  & 	2.082  \\  
			\rowcolor{verylightgray}
			\makecell[c]{Cosine Similarity (NT-Xent)\\ with Multi-positive}   &54.00 & 99.55 & 99.15& 	2.703 \\ 
			\bottomrule[0.8pt]
	\end{tabular}}
 
\end{table}

 \begin{table*}[t]
	\caption{Additional evaluations of the value of chosen $k$ remaining similar samples on MNIST, CIFAR10, and Tiny-ImageNet. \vspace{-4mm}
	}
	\small
	\label{ablation_study_k_remaining_samples2}
	\resizebox{0.98\linewidth}{!}{
		\setlength\tabcolsep{7.5pt}
		\begin{tabular}{ccccccccccccc}
			\toprule[0.8pt]
			\toprule[0.8pt]
			\multirow{2}{*} { \makecell[c]{\textbf{$k$} Value} } & \multicolumn{3}{c}{\textbf{On MNIST}} & \multicolumn{3}{c}{\textbf{On CIFAR10}} & \multicolumn{3}{c}{\textbf{On Tiny-ImageNet}}  &  \multicolumn{3}{c}{\textbf{On CelebA}} \\
			\cmidrule(lr){2-4}
			\cmidrule(lr){5-7}
			\cmidrule(lr){8-10}
			\cmidrule(lr){11-13}
			& MIA (\%)  & RA (\%) & TA (\%) & MIA (\%)  & RA (\%) & TA (\%) & MIA (\%)  & RA (\%) & TA (\%) & MIA (\%)  & RA (\%) & TA (\%)\\
			\midrule
			5 & 62.00 & 99.59 & 99.15    & 59.00 & 99.04 & 81.75   & 54.50 & 81.19 & 56.88  & 54.50 & 96.34 & 95.93\\  
			\rowcolor{verylightgray}
			6 &61.50 & 99.58 & 99.15      & 59.50 & 99.04 & 81.73   & 54.50 & 80.10 & 56.06   & 55.00 & 96.34 & 95.93  \\  
			7 & 61.00 & 99.56 & 98.84    & 59.50 & 99.04 & 81.73   & 54.50 & 80.10 & 56.07   & 55.00 & 96.35 & 95.94  \\  
			\rowcolor{verylightgray}
			8 &62.00 & 99.58 & 99.15    & 59.50 &99.03 & 81.75   & 54.50 &81.20 & 56.89   & 54.00 & 96.47 & 96.17 \\ 
			9 &61.50 & 99.58 & 99.15     & 59.50 & 99.04 & 81.75   & 54.50 & 81.20 & 56.91  & 54.00 & 96.47 & 96.17\\  
			\rowcolor{verylightgray}
			10 &61.50 & 99.58 & 99.15    & 58.50 & 99.05 & 81.75  & 54.50 & 81.19 & 56.90   & 53.00 & 96.48 & 96.19  \\  
			\bottomrule[0.8pt]
	\end{tabular}}
\end{table*}

\subsection{Additional Study: Influence of $k$ Chosen Remaining Similar Positive Samples} \label{add_top_k_ablation}

\noindent
\textbf{Setup.} 
In ManiF-SMC, we choose the top-k remaining most similar samples as the positive set $S_k$, and the anchor is also calculated by the center of the top-k samples. Hence, we test how different top-k values influence our ManiF-SMC method. We randomly select 200 samples for unlearning, and the experimental setting is the same as above. We present the results on MNIST, CIFAR10, TinyImageNet, and CelebA in \Cref{ablation_study_k_remaining_samples2}.

\noindent
\textbf{Results.}
Intuitively, more similar samples in the remaining dataset chosen will increase the model utility preservation during unlearning, because the changes of erasing one sample on the class-centroid can be expressed by $\|x_u - \texttt{c} \| / k$. The results on CelebA in \Cref{ablation_study_k_remaining_samples2} clearly confirm this, higher RA and TA when $k$ value increases. RA increases from 96.34\% ($k=5$) to 96.48\% ($k=10$), and TA climbs from 95.93\% to 96.19\%. However, the unlearning effectiveness drops at the same time when $k$ increases, from 54.50\% ($k=5$) to 53.00\% ($k=10$). Hence, we can claim that a larger $k$ assists to maintain the model utility but mitigates the unlearning effectiveness, which is also proved by other experiments on other datasets in \Cref{ablation_study_k_remaining_samples2}.

\begin{table}[t]
	\caption{Evaluation of checkpoints sampled on the connectivity curve on MNIST. \vspace{-2mm}}
	\label{tab_mnist_connectivity_curve}
	\resizebox{0.95\linewidth}{!}{
		\setlength\tabcolsep{10.5pt}
		\begin{tabular}{cccc}
			\toprule[0.8pt]
			\toprule[0.8pt]
			$t$ & MIA (\%) & RA (\%) & TA (\%) \\
			\midrule
			0.1 & 58.50 & 99.62 & 99.22 \\
			\rowcolor{verylightgray}
			0.3 & 61.00 & 99.63 & 99.21 \\
			0.5 & 62.00 & 99.59 & 99.15 \\
			\rowcolor{verylightgray}
			0.7 & 58.50 & 99.59 & 99.20 \\
			0.9 & 58.00 & 99.62 & 99.21 \\
			\bottomrule[0.8pt]
		\end{tabular}
	}
\end{table}

\subsection{Ablation Study: How does $t$ of SMC Influence ManiF-SMC?} \label{t_eval}

In \Cref{SCM_section}, \Cref{mc_two,eq:w_opt} are applied to train the mode connectivity model to generate the local manifold representation for the adaptive margin calculation. We can train a connectivity curve, such as, $t \in [0,1]$ with step equal to 0.1, i.e., the curve with checkpoints in $\{0, 0.1, 0.2, 0.3, 0.4, 0.5, 0.6, 0.7, 0.8, 0.9, 1\}$. 
In our experiments, we use the checkpoint of $t = 0.5$, which is best for the unlearning effect. We also provide the experiments of the curve in the following \Cref{tab_mnist_connectivity_curve} to demonstrate the results.

\begin{table}[t]
	\caption{Evaluation of different learning models on MNIST (before and after unlearning). \vspace{-4mm}}
	\label{mnist_other_learning_methods}
	\resizebox{0.95\linewidth}{!}{
		\setlength\tabcolsep{6.5pt}
		\begin{tabular}{c|cc|cc|cc}
			\toprule[0.8pt]
			\toprule[0.8pt]
			\multirow{2}{*}{Methods} 
			& \multicolumn{2}{c|}{MIA (\%)} 
			& \multicolumn{2}{c|}{RA (\%)} 
			& \multicolumn{2}{c}{TA (\%)} \\
			\cmidrule(lr){2-3}\cmidrule(lr){4-5}\cmidrule(lr){6-7}
			& Before unl. & After unl. & Before & After & Before & After \\
			\midrule
			Information Bottleneck (IB)      & 51.99 & 61.00 & 99.49 & 99.47 & 98.96 & 98.91 \\
			\rowcolor{verylightgray}
			InfoNCE & 50.50 & 53.50 & 99.39 & 99.40 & 98.91 & 98.83 \\
			\bottomrule[0.8pt]
		\end{tabular}
	}
\end{table}

\begin{table}[t] 
	\caption{Evaluation of unlearning for ViT on MNIST (before and after unlearning). \vspace{-2mm}}
	\label{tab_vit_unlearning1}
	\resizebox{0.98\linewidth}{!}{
		\setlength\tabcolsep{6.0pt}
		\begin{tabular}{c|cc|cc|cc}
			\toprule[0.8pt]
			\toprule[0.8pt]
			\multirow{2}{*}{Model}
			& \multicolumn{2}{c|}{MIA (\%)}
			& \multicolumn{2}{c|}{MSE on remaining set}
			& \multicolumn{2}{c}{MSE on test set} \\
			\cmidrule(lr){2-3}\cmidrule(lr){4-5}\cmidrule(lr){6-7}
			& Before unl. & After unl. & Before & After & Before & After \\
			\midrule
			ViT & 62.00 & 67.00 & 0.0439 & 0.0443 & 0.0442 & 0.0445 \\
			\bottomrule[0.8pt]
		\end{tabular}
	}
\end{table}

\subsection{Application: Scalability to Unlearn Other Models} \label{Scalability_unlearn}

ManiF-SMC unlearning method is also transferable to other representation learning frameworks. We provide the experimental results for InforNCE (contrastive learning) and information bottleneck (IB) (representation learning) in \Cref{mnist_other_learning_methods}.

We also conducted new experiments for Vision Transformer (ViT) on MNIST for the image generative task. Results are provided in \Cref{tab_vit_unlearning1}, indicating the unlearning effectiveness of our method.